\documentclass{article}

\usepackage[preprint]{neurips_2024}

\usepackage[utf8]{inputenc} % allow utf-8 input
\usepackage[T1]{fontenc}    % use 8-bit T1 fonts
\usepackage{hyperref}       % hyperlinks
\usepackage{url}            % simple URL typesetting
\usepackage{booktabs}       % professional-quality tables
\usepackage{amsfonts}       % blackboard math symbols
\usepackage{nicefrac}       % compact symbols for 1/2, etc.
\usepackage{microtype}      % microtypography
\usepackage{xcolor}         % colors
\usepackage{graphicx} 
\usepackage{subfigure}
\usepackage{natbib}
\usepackage{amsmath}
\usepackage{listings}
\usepackage{tcolorbox}
\usepackage{geometry}
\usepackage{cite}
\usepackage{bbm}
\usepackage{float}

\DeclareMathOperator*{\argmax}{arg\,max}

\title{\texttt{Seed-CTS:} Unleashing the Power of Tree Search for Superior Performance in Competitive Coding Tasks}

\author{%
  Hao Wang \\
  ByteDance Seed\\
  \texttt{wanghao.lzy520@bytedance.com} \\
  \And
  Boyi Liu \\
  ByteDance Seed \\
  \texttt{boyi.liu01@bytedance.com} \\
  \And
  Yufeng Zhang \\
  ByteDance Seed \\
  \texttt{yufeng.zhang@bytedance.com} \\
  \And
  Jie Chen\thanks{Corresponding author and project leader.} \\
  ByteDance Seed \\
  \texttt{chenjiexjtu@gmail.com} \\
}

\begin{document}

\maketitle

\begin{abstract}
Competition-level code generation tasks pose significant challenges for current state-of-the-art large language models (LLMs). For example, on the LiveCodeBench-Hard dataset, models such as O1-Mini and O1-Preview achieve pass@1 rates of only 0.366 and 0.143, respectively. While tree search techniques have proven effective in domains like mathematics and general coding, their potential in competition-level code generation remains under-explored. In this work, we propose a novel token-level tree search method specifically designed for code generation. Leveraging Qwen2.5-Coder-32B-Instruct, our approach achieves a pass rate of \textbf{0.305} on LiveCodeBench-Hard, surpassing the pass@100 performance of GPT4o-0513 (0.245). Furthermore, by integrating Chain-of-Thought (CoT) prompting, we improve our method's performance to \textbf{0.351}, approaching O1-Mini’s pass@1 rate. To ensure reproducibility, we report the average number of generations required per problem by our tree search method on the test set. Our findings underscore the potential of tree search to significantly enhance performance on competition-level code generation tasks. This opens up new possibilities for large-scale synthesis of challenging code problems supervised fine-tuning (SFT) data, advancing competition-level code generation tasks.
\end{abstract}

% \begin{figure}[!tbp]
%   \centering
%     \subfigure[]{
%         \includegraphics[width=0.95\textwidth]{figures/LiveCodeBench-Hard_DS6B7_1.png}
%         \label{pass_100}
%     }
%     \subfigure[]{
%         \includegraphics[width=0.95\textwidth]{figures/LiveCodeBench-Hard_DS6B7_2.png}
%         \label{pass_k}
%     }
%     \caption{(a) We compare the pass rates of MCTS under different values of max\_rollouts with the pass@100 scores of sota models on LiveCodeBench-Hard. (b) We compare the pass rates of MCTS under different values of max\_rollouts with the pass@1, pass@10, and pass@100 scores of Qwen2.5-72B-Instruct-api on LiveCodeBench-Hard.}
%     \label{LiveCodeBench-Hard}
% \end{figure}

\section{Introduction}

Competition-level code generation tasks present a unique set of challenges for large language models (LLMs). These tasks require models to not only comprehend complex problem statements but also generate executable code that adheres to logical and syntactical constraints. While existing state-of-the-art LLMs have achieved remarkable success in general-purpose programming benchmarks, their performance on competitive programming datasets, such as LiveCodeBench-Hard~\citep{naman2024livecodebench}, remains far from satisfactory. For example, recent models like O1-Mini and O1-Preview exhibit pass@1 rates of only 0.366 and 0.143, respectively. This performance gap highlights the need for novel methodologies to enhance model capabilities in solving these challenging tasks.

Recent research has demonstrated the potential of tree search techniques in reasoning tasks like mathematics and general programming. However, their application to competition-level code generation remains under-explored. Existing approaches primarily rely on large-scale proprietary LLMs within tree search frameworks, overlooking the possibility that smaller, open-source models—when paired with an effective search strategy—could achieve superior results. Moreover, while data augmentation through techniques such as distillation from stronger LLMs has been widely used, generating solutions directly from the target model itself offers the potential for higher-quality supervised fine-tuning (SFT) data, as these solutions are directly generated by the target model, ensuring consistency with its inherent capabilities and output characteristics.

\begin{figure}[!tbp]
    \centering
    \includegraphics[width=0.8\textwidth]{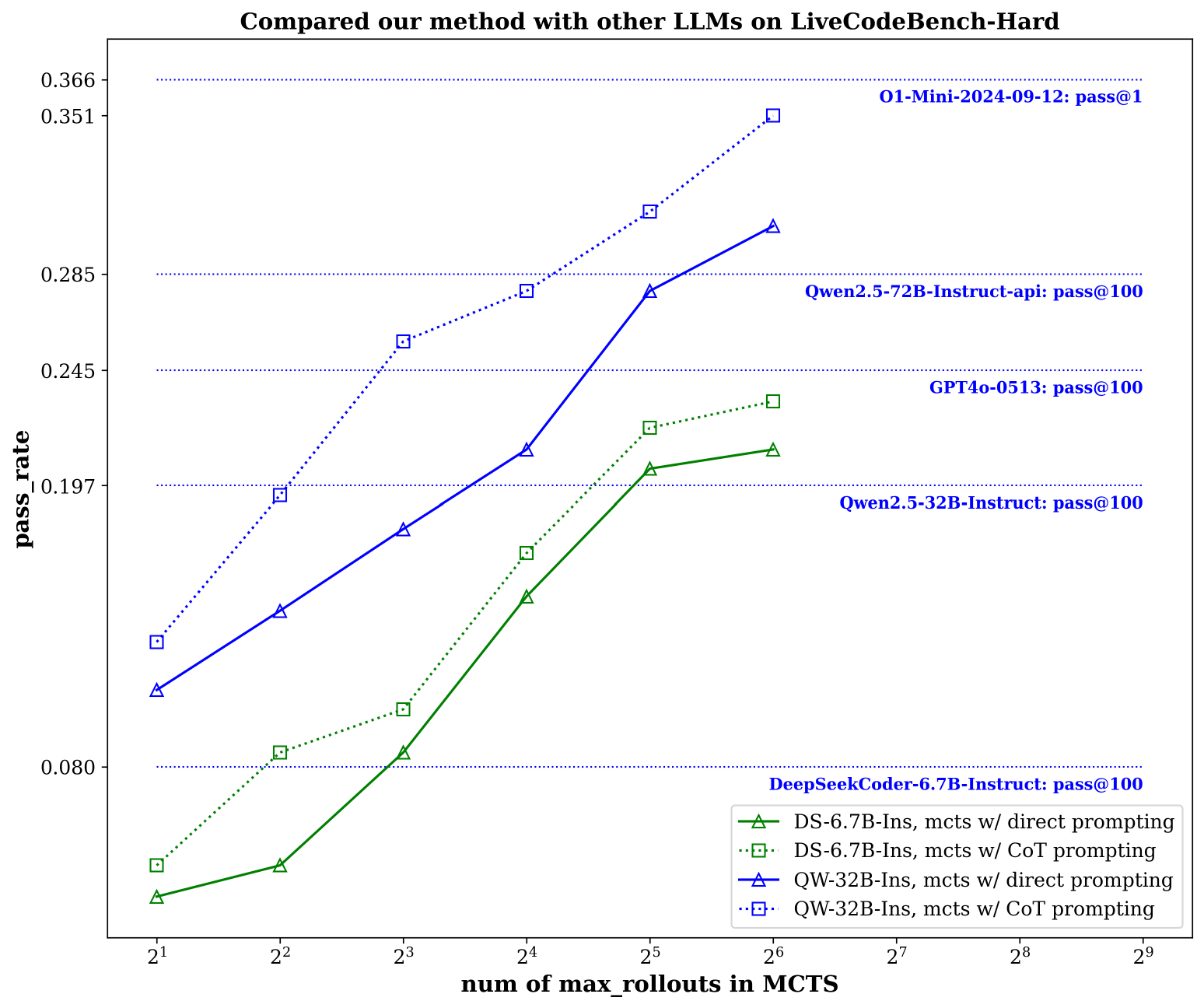}
    \caption{Pass rates of MCTS with DeepSeekCoder-6.7B-Instruct and Qwen2.5-32B-Instruct on LiveCodeBench-Hard: each model surpasses it's own pass@100 rates at $\text{max\_rollouts}=8$. Notably, MCTS with Qwen2.5-32B-Instruct and $\text{max\_rollouts}=16$ outperforms pass@100 of both Qwen2.5-72B-Instruct and GPT4o-0513. In addition, when combined with CoT prompting, MCTS with Qwen2.5-32B-Instruct achieves a pass rate of 0.351 nearing O1-Mini's pass@1 rate of 0.366.}
    \label{fig:hard-intro}
\end{figure}

In this work, we propose a novel token-level Monte Carlo Tree Search (MCTS) method tailored specifically for competition-level code generation. Leveraging the open-source Qwen2.5-Coder-32B-Instruct model, as shown in Figure~\ref{fig:hard-intro}, our approach achieves a pass rate of \textbf{0.305} on LiveCodeBench-Hard, surpassing the pass@100 performance of GPT4o-0513 (0.245). By incorporating Chain-of-Thought (CoT) prompting, our method further improves to \textbf{0.351}, approaching O1-Mini's pass@1 rate. These results demonstrate that our method not only enhances the ability of models to solve previously unsolvable problems but also provides high-quality outputs that can be directly used to synthesize new SFT data. Compared to distillation-based approaches that rely on external models, our framework allows for a more intrinsic and effective alignment of the data with the target model's capabilities.

We summarize our contributions as follows:

\begin{itemize}
    \item \textbf{A novel token-level MCTS framework with Cot Prompting:} We propose a token-level tree search framework that combines MCTS with CoT prompting, enabling iterative refinement of both reasoning and code generation for competition-level tasks.
    \item \textbf{Enhancing open-source models for competitive programming tasks:} We demonstrate that open-source models like Qwen2.5-Coder-32B-Instruct can achieve substantial performance improvements on competitive programming datasets when paired with our method. This showcases the potential to elevate the capabilities of open-source models to rival or surpass proprietary counterparts in challenging domains.
    \item \textbf{Comprehensive experimental analysis:} We demonstrate the efficacy of the proposed framework while reporting efficiency metrics such as average generations per problem to ensure reproducibility and fairness.
\end{itemize}
% \begin{figure}
%   \begin{minipage}[c]{0.65\textwidth}
%     \includegraphics[width=\textwidth]{figures/hard-intro.pdf}
%   \end{minipage}\hfill
%   \begin{minipage}[c]{0.315\textwidth}
%     \caption{Pass rates of MCTS with DeepSeekCoder-6.7B-Instruct and Qwen2.5-32B-Instruct on LiveCodeBench-Hard: each model surpasses it's own pass@100 rates at $\text{max\_rollouts}=8$. Notably, MCTS with Qwen2.5-32B-Instruct and $\text{max\_rollouts}=16$ outperforms pass@100 of both Qwen2.5-72B-Instruct and GPT4o-0513. In addition, when combined with CoT prompting, MCTS with Qwen2.5-32B-Instruct achieves a pass rate of 0.351 nearing O1-Mini's pass@1 rate of 0.366.
%     } \label{fig:hard-intro}
%   \end{minipage}
% \end{figure}

\section{Related Work}

\subsection{LLMs for Code Generation}
Large language models (LLMs), with their powerful reasoning capabilities, have been widely adopted in code-related research and applications. The primary approach to building code LLMs involves pre-training or fine-tuning them on large code datasets, such as CodeX~\citep{chen2021evaluating}, AlphaCode  \citep{li2022competition}, WizardCoder~\citep{luo2023wizardcoder}, CodeGeeX~\citep{zheng2023codegeex}, Starcoder~\citep{li2023starcoder} and Code LLama~\citep{roziere2023code}. Foundation models, like GPT-4~\citep{achiam2023gpt} and Claude\footnote{\url{https://claude.ai/}}, exhibit remarkable code generation capabilities despite lacking additional fine-tuning on code-specific data. Additionally, building upon the robust planning capabilities~\citep{yao2022react} and reflection mechanisms~\citep{shinn2024reflexion} of LLMs, LLM-powered autonomous agents have shown significant potential in advancing automated code generation~\citep{huang2023agentcoder, hong2023metagpt, wang2024executable, zhang2024autocoderover}. For example, Agentcoder~\citep{huang2023agentcoder} proposes a multi-agent framework that includes programmer agents, test designer agents, and test execution agents to collaboratively generate and test code, MetaGPT~\citep{hong2023metagpt} imitates the main roles in software companies in the real world, using different AI agents to play and ultimately produce a project. 

\subsection{Prompt Engineering}
Designing effective prompts to seamlessly communicate with LLMs to fully harness their full potential can significantly improve LLMs performance without additional training. Some representative technologies of prompt engineering include Chain-of-Thought (CoT)~\citep{wei2022chain}, Self-Consistency~\citep{wang2022self}, Tree-of-Thought (ToT)~\citep{yao2024tree}, Reasoning via Planning (RAP)~\citep{hao2023reasoning} and Self-Refine~\citep{madaan2024self}. This technique can be directly applied in LLM for iterative and self improving (refining) code generation. For instance, CodeCoT~\citep{huang2023codecot} integrates chain-of-thought reasoning with a self-examination process, iteratively refining code based on execution feedback to ensure both logical correctness and syntactic validity. Self-planning~\citep{jiang2024self} enhances code generation by using LLMs to first plan solution steps from intent and then generate code step-by-step. Self-debugging~\citep{chen2023teaching}, an LLM is prompted to iteratively refine code predictions by utilizing feedback from explanations and execution results to identify and fix errors.

\subsection{Monte Carlo Tree Search (MCTS) for Reasoning}
\citet{chen2021evaluating} showed that repeated sampling can produce correct code solutions, suggesting the answer lies within the LLMs' output space with notable probability, motivating the use of tree search for efficient exploration~\citep{li2024rethinkmcts, qi2024mutual, wang2024q, hui2024rot}. PG-TD~\citep{zhang2023planning} introduces a planning-guided Transformer decoding algorithm that uses MCTS and test-case evaluation to iteratively refine code generation, \citet{zhang2024rest} proposed ReST-MCTS*, a method that integrates process reward guidance with tree search to infer high-quality reasoning traces and per-step rewards, enabling more effective self-training of policy. Another common method is LATS~\citep{zhou2023language}, which leverages LLMs as agents, value functions, and optimizers, incorporating MCTS to enhance decision-making through external feedback and experience. PlanSearch~\citep{wang2024planning} improves code generation by searching over diverse natural language plans instead of directly over code.

\section{Method}

\subsection{Preliminary}

Neural code generation aims to automatically transform natural language descriptions into executable source code through large language models (LLMs). Here we provide a formal definition of the code generation task.

Let $D$ represent a natural language description of a programming task, which may include problem statements, requirements, and additional programming context such as function signatures or assertions. The code generation task can be formalized as learning a model $\pi_\theta$ parameterized by $\theta$ that generates code solution $C$ given description $D$:
\begin{equation}
    C \sim \pi_\theta(\cdot \,|\, D).
\end{equation}

To evaluate the correctness of generated code, we define a test suite $T = \{(x_i, y_i)\}_{i=1}^n$ where each test case consists of an input $x_i$ and its expected output $y_i$. The test suite is typically divided into two subsets
\begin{equation}
T = T_{\text{pub}} \cup T_{\text{priv}},    
\end{equation}
where $T_{\text{pub}}$ represents public test cases visible during development, and $T_{\text{priv}}$ represents private test cases held out for evaluation.

Given a code solution $C$, we define an execution function $\text{Exec}(C, x)$ that returns the output of running $C$ on input $x$. The correctness of $C$ can then be measured by comparing its outputs against the expected outputs across all test cases:
\begin{equation}
    \text{Correct}(C\,|\,D, T) = \frac{1}{|T|}\sum_{(x,y) \in T} \mathbbm{1}(\text{Exec}(C, x) = y),
\end{equation}
where $\mathbbm{1}(\cdot)$ is the indicator function.% that returns 1 if the condition is true and 0 otherwise.

The objective of code generation is to find model parameters $\theta^*$ that maximize the expected correctness across a distribution of programming tasks:
\begin{equation}
    \theta^* = \argmax_\theta \mathbb{E}_{D,T}[\text{Correct}(\pi_\theta(D))].
\end{equation}

\subsection{MCTS with CoT Prompting}
The proposed method is motivated by the desire to integrate Chain-of-Thought (CoT) reasoning with Monte Carlo Tree Search (MCTS). Specifically, the approach enables LLMs to first generate intermediate reasoning steps, followed by code generation. Through iterative refinement and optimization of both the reasoning and code components via MCTS, the method aims to enhance the model's performance on challenging competition-level code generation tasks. Next, we will provide a detailed description of our methods in each key component of MCTS.

\noindent{\bf CoT Prompting.} To improve the performance of LLMs on challenging competition-level code generation tasks, we introduce a structured CoT prompting methodology, which guides the model through a two-step reasoning process \textbf{planning} and \textbf{coding} to ensure logical and syntactically correct outputs. MCTS iteratively refines and optimizes both the planning and coding stages, improving performance in complex code generation tasks. The prompt explicitly instructs the model to:
\begin{itemize}
    \item \textbf{Solution Planning:} Analyze the problem specification and create a detailed step-by-step plan. This step includes outlining the problem-solving logic, choosing appropriate data structures, and determining the functions required for implementation.
    \item \textbf{Code Generation:} Based on the detailed plan, write Python code adhering to coding standards and ensuring proper syntax.
\end{itemize}

Here is an example:

\definecolor{codegray}{rgb}{0.5,0.5,0.5}
\definecolor{codepurple}{rgb}{0.58,0,0.82}
\definecolor{backcolour}{rgb}{0.95,0.95,0.92}

\lstdefinestyle{mystyle}{
    backgroundcolor=\color{backcolour},
    commentstyle=\color{codegray},
    keywordstyle=\color{blue},
    numberstyle=\tiny\color{codegray},
    stringstyle=\color{codepurple},
    basicstyle=\ttfamily\footnotesize,
    breakatwhitespace=false,
    breaklines=true,
    captionpos=b,
    keepspaces=true,
    numbers=left,
    numbersep=5pt,
    showspaces=false,
    showstringspaces=false,
    showtabs=false,
    tabsize=2
}

\lstset{style=mystyle}

\begin{tcolorbox}[colback=white, colframe=blue!75, arc=3mm, boxrule=1mm, width=\textwidth, title=Cot Prompting Example]
\textbf{Instruction:} As an AI language model, you are tasked with generating Python code based on given problem specifications. To accomplish this, you need to follow a two-step process:

1. **Planning**: Analyze the problem and create a detailed plan that outlines the steps needed to solve the problem. This plan should include the logic, data structures, and Python functions you'll use.

2. **Coding**: Based on the plan, write the Python code that solves the problem. Make sure your code adheres to the Python coding standards and uses the correct syntax.

\textbf{Problem:} You are given a 0-indexed integer array \texttt{nums}. The teacher wants to form a group of students with maximal strength. The strength of a group is defined as the product of all the elements in the group. Return the maximum strength of a group the teacher can create.

\textbf{Plan:}

<startofplan>

Step 1: Import the combinations function from itertools module.

Step 2: Define a class Solution with a method maxStrength that takes a list of integers as input.

......

Step 8: After all combinations are processed, return max\_strength.

<endofplan>

\textbf{Python Code:}

\begin{lstlisting}[language=Python]
from itertools import combinations

class Solution:
    def maxStrength(self, nums: List[int]) -> int:
        if len(nums) == 1:
            return nums[0]
        max_strength = float('-inf')
        for i in range(1, len(nums)+1):
            for comb in combinations(nums, i):
                strength = 1
                for num in comb:
                    strength *= num
                max_strength = max(max_strength, strength)
        return max_strength
\end{lstlisting}
\end{tcolorbox}

\noindent{\bf Selection.} The selection phase in MCTS strives to balance exploration and exploitation by selecting actions that are most likely to yield beneficial results. At the selection stage, the algorithm starts from the root node $s_0$ and traverses the tree until it reaches a leaf node. Our method use a token-level MCTS so that each state $s$ represents a candidate token. At each node $s$, the action $a \in \mathcal{A}(s)$, where $\mathcal{A}(s)$ denotes the set of available actions in state $s$ taken by the LLM $\pi$, is chosen by maximizing the P-UCB score:
\begin{align}
\text{P-UCB}(s, a) &= Q(s, a) + \beta(s) \cdot p(a \,|\, s) \cdot \frac{\sqrt{\ln N(s)}}{1 + N(s, a)},\\
\beta(s) &= \log\left(\frac{{N(s)} + c_\text{base} + 1}{c_\text{base}}\right) + c.
\end{align}
Here:
\begin{itemize}
\item $Q(s, a)$ represents the average reward (defined in \textbf{Simulation}) of action $a$ at state $s$.
\item $N(s)$ is the total number of visits to state $s$.
\item $N(s, a)$ is the number of times action $a$ has been taken from state $s$.
\item $p(a\,|\, s)$ is the prior probability of action $a$ at $s$, proposed by the LLM $\pi$.
\item $c_{\text{base}}$ and $c$ are hyperparameters that balance exploration and exploitation.
\end{itemize}
This formula combines three essential components: exploitation through $Q(s, a)$, exploration driven by $\sqrt{{\ln N(s)}/(1 + N(s, a))}$, and prior guidance from $P(s, a)$. The algorithm iteratively applies this criterion until it encounters a leaf node $s_L$, defined as a state either not fully expanded or terminal.

\noindent{\bf Expansion.} When the selection process reaches a node $s$ in the search tree, the expansion phase creates new child nodes by considering potential next tokens. Unlike standard MCTS which might randomly sample actions, we leverage the LLM's predictions to guide expansion:

Given the current state $s$, we obtain the k most probable next tokens using the TOP-K function:
\begin{equation}
\mathcal{T}_k(s) = \text{TOP\_K}(s, k),
\end{equation}
where $k$ is a hyperparameter that limits the maximum number of children per node, and $\mathcal{T}_k(s)$ returns the set of $k$ most likely next tokens according to the LLM's probability distribution.

For the new child node $s'$, the visit count $N(s', a')$ and the average reward $Q(s', a')$ for all $a' \in \mathcal{A}(s')$ are initialized to zero:
\begin{equation}
N(s', a') = 0, \quad Q(s', a') = 0, \qquad \forall a' \in \mathcal{A}(s').
\end{equation}
The use of priors $p(s'\,|\, a')$ derived from the policy $\pi$ enables the tree to bias future expansions toward promising regions of the search space.

\noindent{\bf Simulation.} Once a new node $s'$ is added to the tree, the algorithm estimates its value through a simulation, also called a rollout. Starting from $s'$, actions are sampled according to the policy $\pi$ until a terminal state $s_T$ is reached or a predefined depth limit $d_\text{max}$ is exceeded. In this paper, we use two methods to estimate the quality for the state $s'$, we call it as hard reward (HR) and partial reward (PR). Normally, we use all public test cases to validate the generated code. HR supposes that if everything passes, the code is considered correct, and if there exist errors, it is incorrect. Assume that $T$ is the set of all test cases, the HR can be formalized as:
\begin{equation}
    R_{s'}^{\text{HR}} = 
\begin{cases} 
1, & \text{if } \mathbbm{1}(\text{Exec}(C, x) = y) = 1,  \forall (x,y) \in T; \\
0, & \text{otherwise}.
\end{cases}
\end{equation}

However, when addressing challenging coding problems, the distinction between partial success and complete failure is critical. Accordingly, our method leverages the pass rate on the test set as the reward signal, denoted as PR:
\begin{equation}
R_{s'}^{\text{PR}} = \frac{1}{|T|}\sum_{(x,y) \in T} \mathbbm{1}(\text{Exec}(C, x) = y).
\end{equation}
\noindent{\bf Backpropagation.} The backpropagation stage updates the statistics of all nodes along the path from the newly expanded node $s'$ back to the root $s_0$. For each node-action pair $(s,a)$ on the path, the visit count $N(s,a)$ and the average reward $Q(s,a)$ are updated as follows:
\begin{align}
N(s, a) &\leftarrow N(s, a) + 1, \\
Q(s, a) &\leftarrow \frac{(N(s, a) - 1) \cdot Q(s, a) + R(s')}{N(s, a)}.
\end{align}
These updates propagate the simulation result $R(s')$ upward, refining the action-value estimates $Q(s,a)$ and balancing the contributions of exploration and exploitation.

\begin{figure}[t!]
  \centering
  \subfigure[]{
        \includegraphics[width=0.48\textwidth]{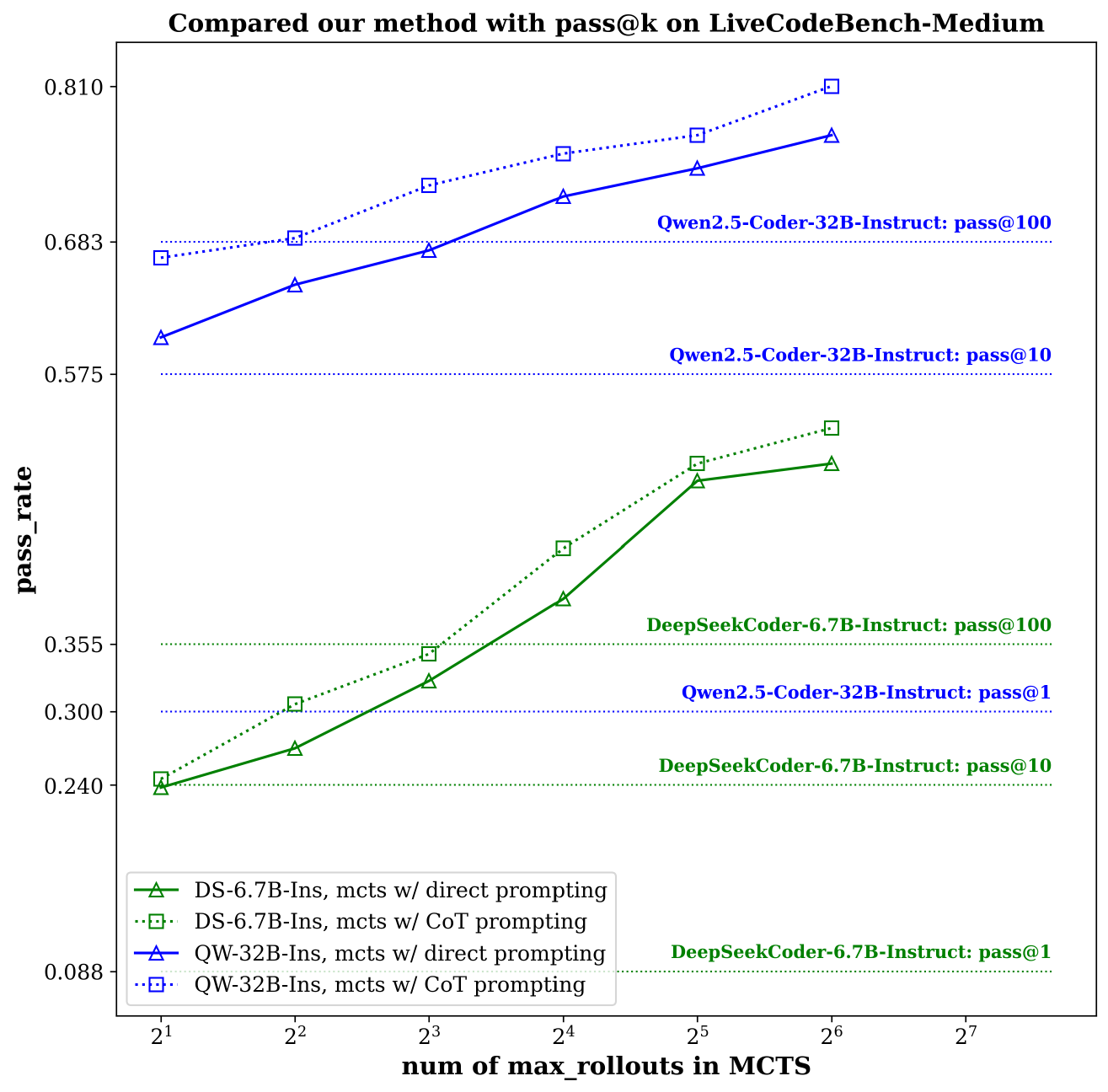}
        \label{LiveCodeBench-Medium_pass_k}
    }
    \subfigure[]{
        \includegraphics[width=0.48\textwidth]{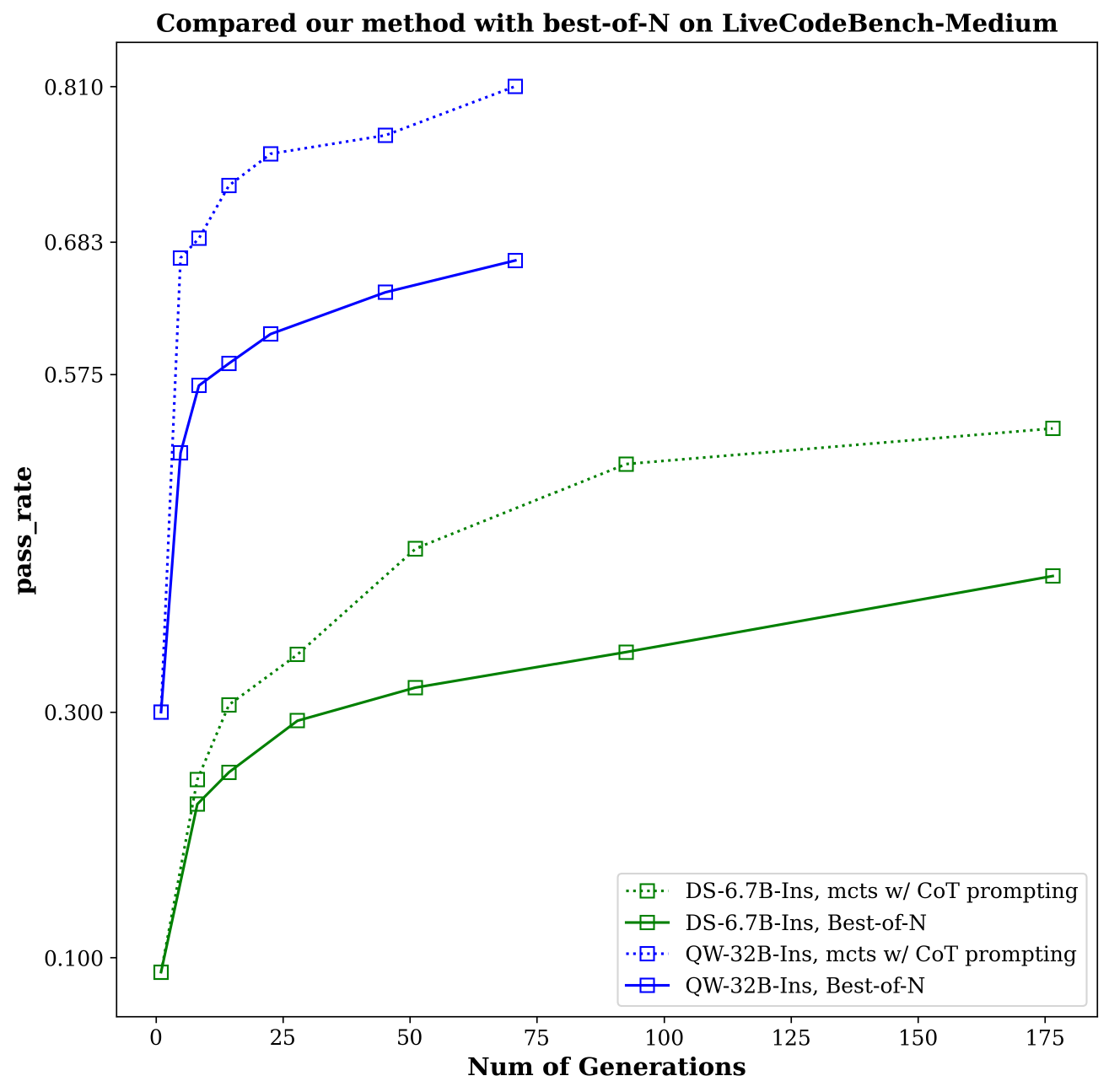}
        \label{LiveCodeBench-Medium_pass_100}
    }
      \subfigure[]{
        \includegraphics[width=0.48\textwidth]{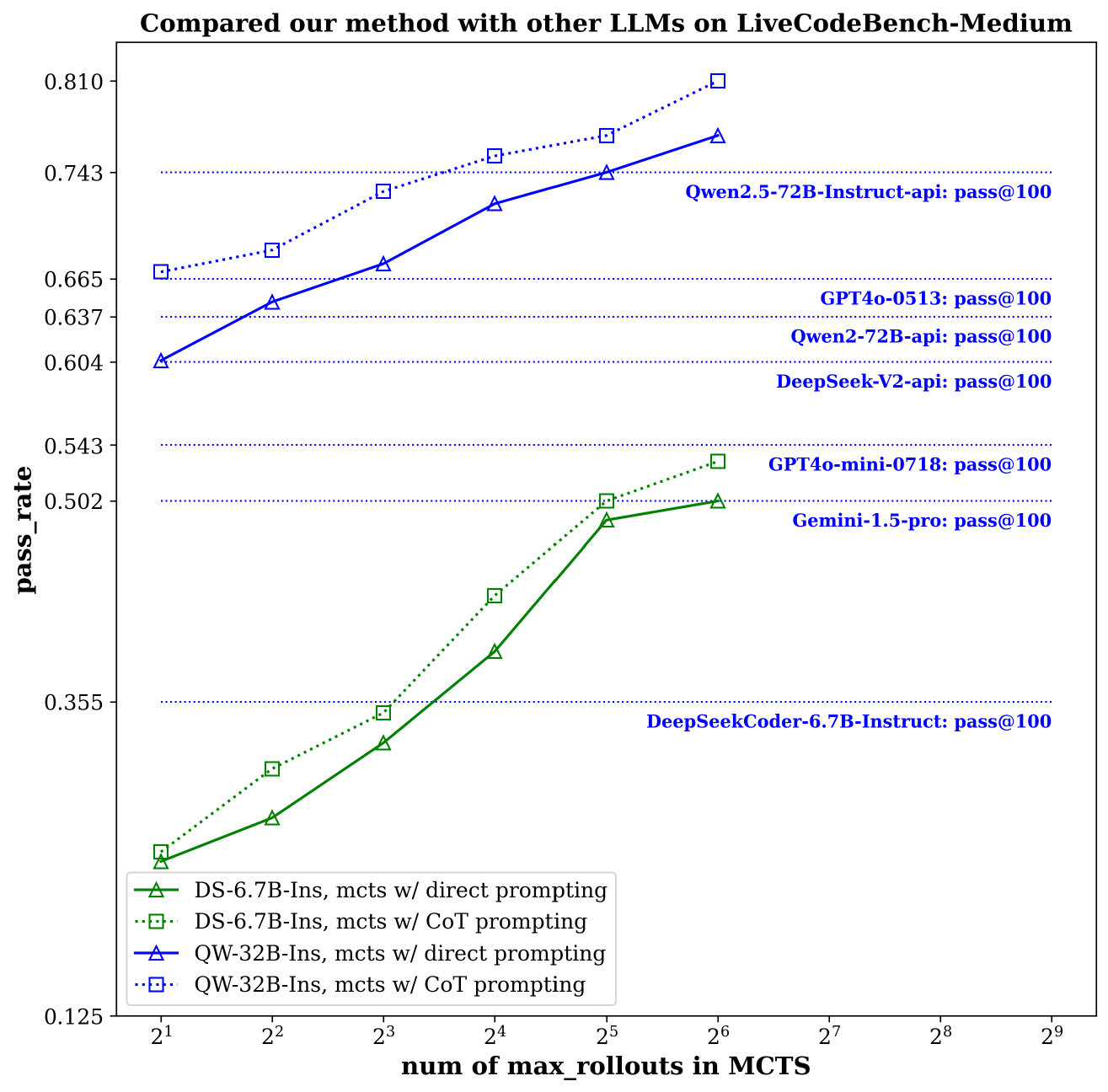}
        \label{LiveCodeBench-Medium_pass_k-2}
    }
    \subfigure[]{
        \includegraphics[width=0.48\textwidth]{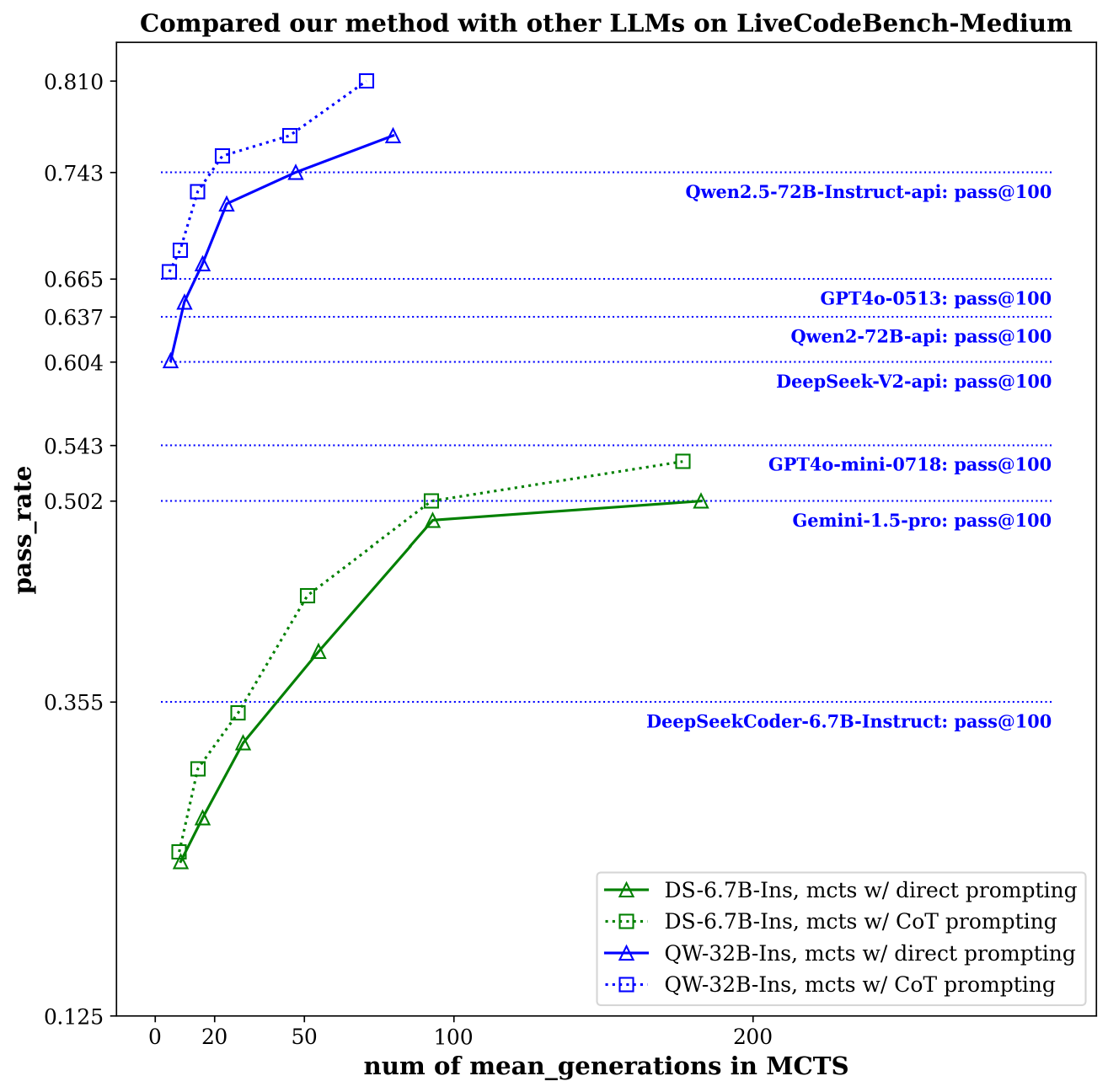}
        \label{LiveCodeBench-Medium_pass_100-2}
    }
    \caption{Results on LiveCodeBench-Medium: (a) Comparison of the pass rates of MCTS with different max\_rollouts against the pass@1, pass@10, and pass@100 rates of Qwen2.5-72B-Instruct-api. (b) Comparison of pass rates of MCTS against pass@k with k selected as the mean number of generations in the corresponding MCTS run. (c)-(d) Comparison of the pass rates of MCTS under different max\_rollouts against the pass@100 rates of sota models.}
    \label{LiveCodeBench-Medium}
\end{figure}

\section{Experiment}
We use LiveCodeBench~\citep{naman2024livecodebench} as the test dataset, which comprises 659 problems collected between May 1, 2023, and September 1, 2024. The dataset categorizes problems into three difficulty levels: easy, medium, and hard, with 245 medium-level problems and 151 hard-level problems. We expand 5 child nodes in the expansion phase. In the simulation phase, we set the temperature = 0.7, top\_p = 0.8 and repetition\_penalty = 1.05 to balance randomness, diversity, and coherence. %Considering that state-of-the-art models have already achieved high pass rates on easy problems—such as Claude-3.5-Sonnet-20240620, which has a pass@1 rate of 86.9\%—our study focuses on the medium and hard subsets.
Given that state-of-the-art models, such as Claude-3.5-Sonnet-20240620, have achieved a pass@1 rate of 0.869 on the easy subset, evaluating the impact of alternative inference techniques on these problems would likely be uninformative. Therefore, our study focuses on the medium and hard subsets, where there is greater scope for improvement and more meaningful differences in inference performance.

To validate the model-agnostic nature of the proposed MCTS approach, we employ four generative models, DeepSeekCoder-6.7B-Instruct, Qwen2.5-7B-Instruct, Qwen2.5-14B-Instruct, and Qwen2.5-32B-Instruct, in our experiment.

\subsection{MCTS on LiveCodeBench}
In this section, we evaluate the performance of MCTS across several different generating models on LiveCodeBench-Medium and LiveCodeBench-Hard. For better readability, we present plots in what follows and defer detailed tables to Appendix \ref{appendix:table}.

\noindent{\bf Medium Level.} We begin by comparing the performance of MCTS against the pass@k rates for the same generating models. Figure~\ref{LiveCodeBench-Medium_pass_k} illustrates the performance of MCTS with DeepSeekCoder-6.7B-Instruct and Qwen2.5-32B-Instruct as generating models, compared against their pass@k rates on LiveCodeBench-Medium. In addition, Figure~\ref{LiveCodeBench-Medium_pass_100} illustrates the performance variation of MCTS and Best-of-N (or pass@k) as the number of number of generations increases. Across all tested configurations, MCTS consistently outperforms the Best-of-N baselines, which demonstrates its effectiveness in leveraging the same underlying model to achieve superior results.

In the second part of our analysis, Figure~\ref{LiveCodeBench-Medium_pass_k-2} compares the performance of MCTS with smaller models (DeepSeekCoder-6.7B-Instruct and Qwen2.5-32B-Instruct) against the pass@100 rates of much larger and more capable models. Key results include: when $\text{max\_rollouts}=32$, the pass@100 rate of MCTS with DeepSeekCoder-6.7B-Instruct reaches 0.488, comparable to the pass@100 rate of Gemini-1.5-pro at 0.502. With $\text{max\_rollouts}=64$, MCTS with Qwen2.5-32B-Instruct achieves a pass rate of 0.770, surpassing the pass@100 rates of much larger models such as Qwen2.5-72B-Instruct-api and GPT4o-0513. Further insights are provided in Figure~\ref{LiveCodeBench-Medium_pass_100-2}, where the x-axis represents the mean number of generations. The results demonstrate that the proposed MCTS approach achieves higher pass rates with fewer sampling attempts, emphasizing its efficiency and effectiveness.

% \begin{figure}[!tbp]
%   \centering
%   \subfigure[]{
%         \includegraphics[width=0.48\textwidth]{figures/LiveCodeBench-Medium_DS6B7_1.png}
%         \label{LiveCodeBench-Medium_pass_k-2}
%     }
%     \subfigure[]{
%         \includegraphics[width=0.48\textwidth]{figures/LiveCodeBench-Medium_DS6B7_1_generations.png}
%         \label{LiveCodeBench-Medium_pass_100-2}
%     }
%     \caption{(a) We compare the pass rates of MCTS under different values of max\_rollouts with the pass@1, pass@10, and pass@100 scores of Qwen2.5-72B-Instruct-api on LiveCodeBench-Medium. (b) We compare the pass rates of MCTS under different values of max\_rollouts with the pass@100 scores of sota models on LiveCodeBench-Medium.}
%     \label{LiveCodeBench-Medium-2}
% \end{figure}

% We defer the detailed experiment results to Table~\ref{baseline-pass@k-LiveCodeBench-Medium} and Table~\ref{baseline-MCTS-LiveCodeBench-Medium}.
\begin{figure}[t!]
  \centering
  \subfigure[]{
        \includegraphics[width=0.48\textwidth]{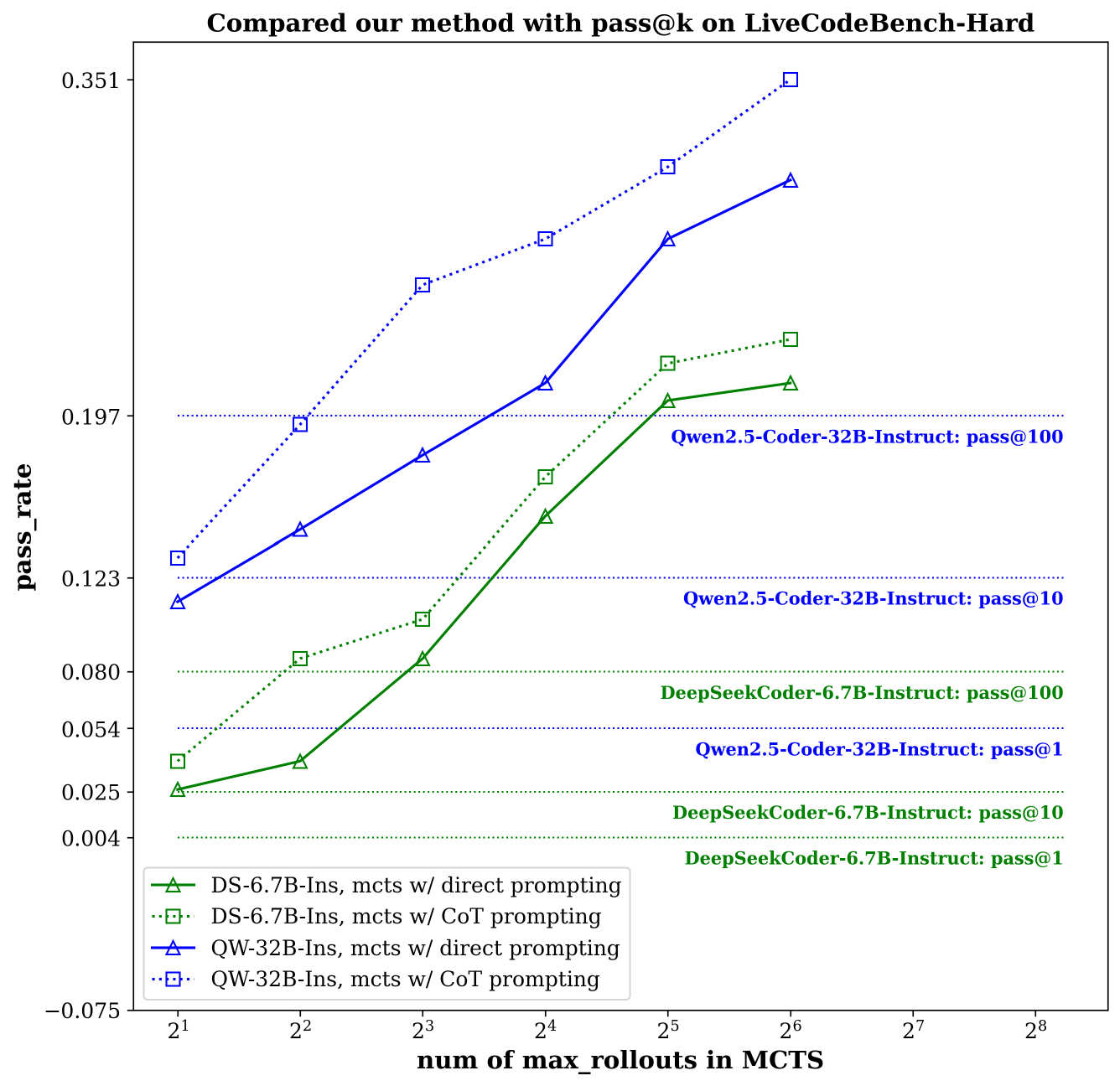}
        \label{LiveCodeBench-Hard-self}
    }
    \subfigure[]{
        \includegraphics[width=0.48\textwidth]{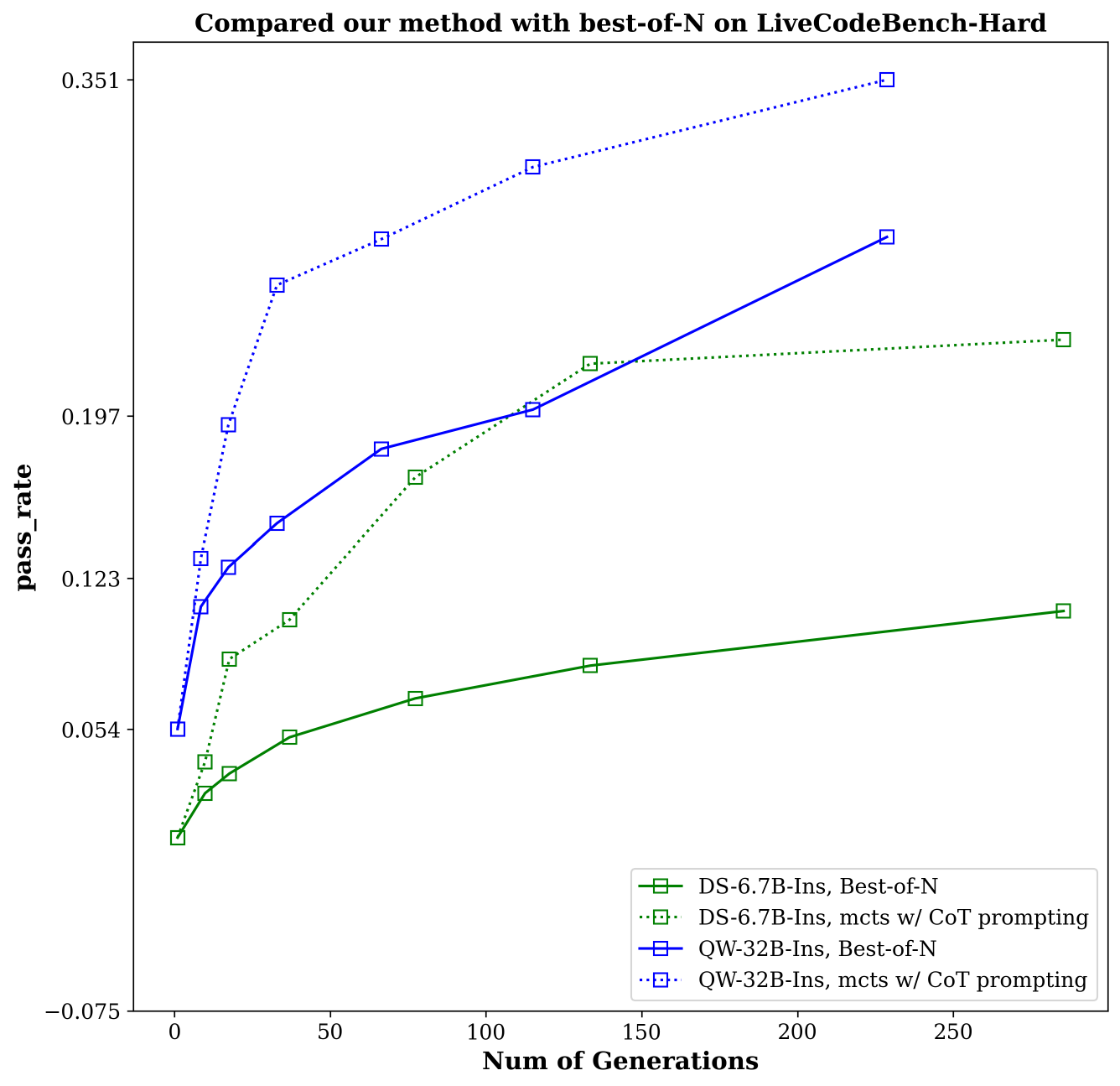}
        \label{LiveCodeBench-Hard-bon}
    }
  \subfigure[]{
        \includegraphics[width=0.48\textwidth]{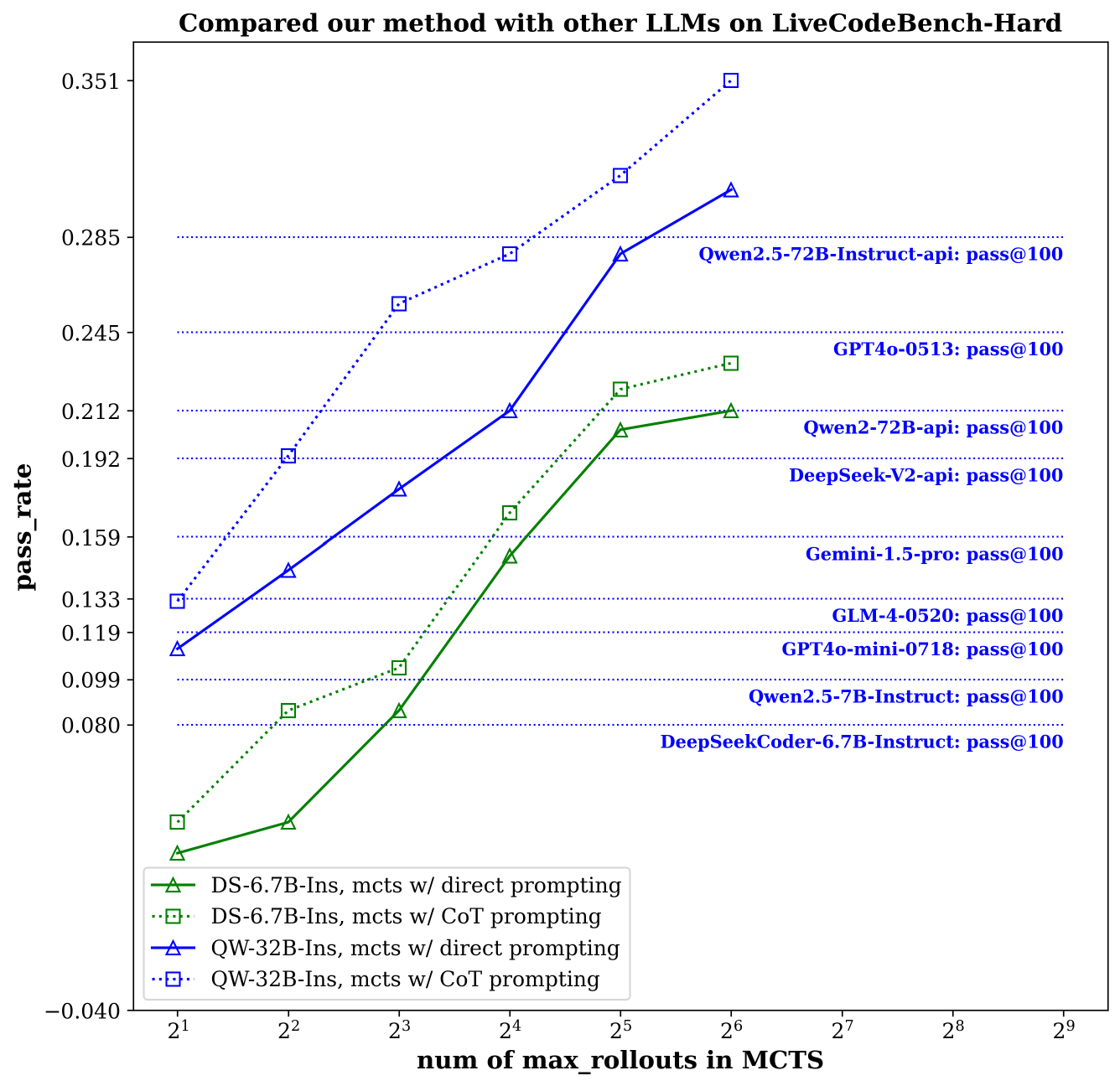}
        \label{LiveCodeBench-Hard-sota}
    }
    \subfigure[]{
        \includegraphics[width=0.48\textwidth]{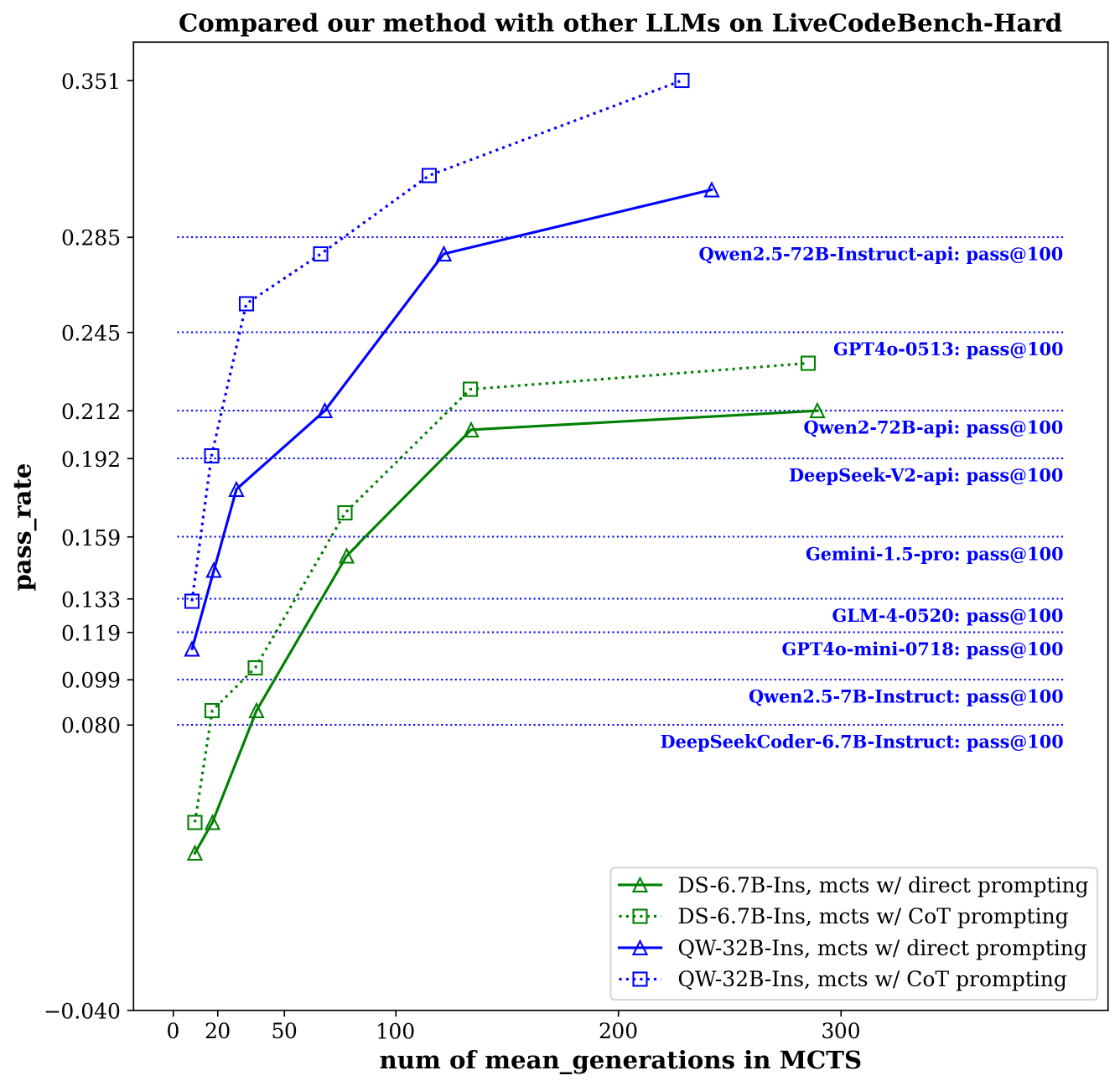}
        \label{LiveCodeBench-Hard-gen}
    }
    \caption{Results on LiveCodeBench-Hard: (a) Comparison of the pass rates of MCTS with different max\_rollouts against the pass@1, pass@10, and pass@100 rates of Qwen2.5-72B-Instruct-api. (b) Comparison of pass rates of MCTS against pass@k with k selected as the mean number of generations in the corresponding MCTS run. (c)-(d) Comparison of the pass rates of MCTS under different max\_rollouts against the pass@100 rates of sota models.}
    \label{LiveCodeBench-Hard}
\end{figure}

\noindent{\bf Hard Level.} The LiveCodeBench-Hard subset poses significantly greater challenges compared to LiveCodeBench-Medium. Notably, Qwen2.5-72B-Instruct-api achieves the highest pass@1 rate of only 0.087, with a pass@100 rate of 0.285. Both DeepSeekCoder-6.7B-Instruct and Qwen2.5-7B-Instruct exhibit pass@100 rates below 10, underscoring the difficulty of this subset. For detailed pass@k rates of state-of-the-art models on LiveCodeBench-Hard, we refer readers to Table~\ref{baseline-LiveCodeBench-Hard-passk} in Appendix \ref{appendix:table}.

Figure~\ref{LiveCodeBench-Hard-self} illustrates the pass@100 rates of various models on LiveCodeBench-Hard, along with the performance variation of MCTS using DeepSeekCoder-6.7B-Instruct and Qwen2.5-32B-Instruct as generating models as the number of max\_rollouts increases. Notably, as shown in Figure~\ref{LiveCodeBench-Hard-bon}, MCTS with DeepSeekCoder-6.7B-Instruct begins to surpass the Best-of-N performance of Qwen2.5-32B-Instruct when $\text{max\_rollouts} \geq 16$. These advantages of MCTS become even more compelling as the difficulty of the test dataset increases.

\begin{figure}[t!]
    \centering
    \includegraphics[width=0.8\textwidth]{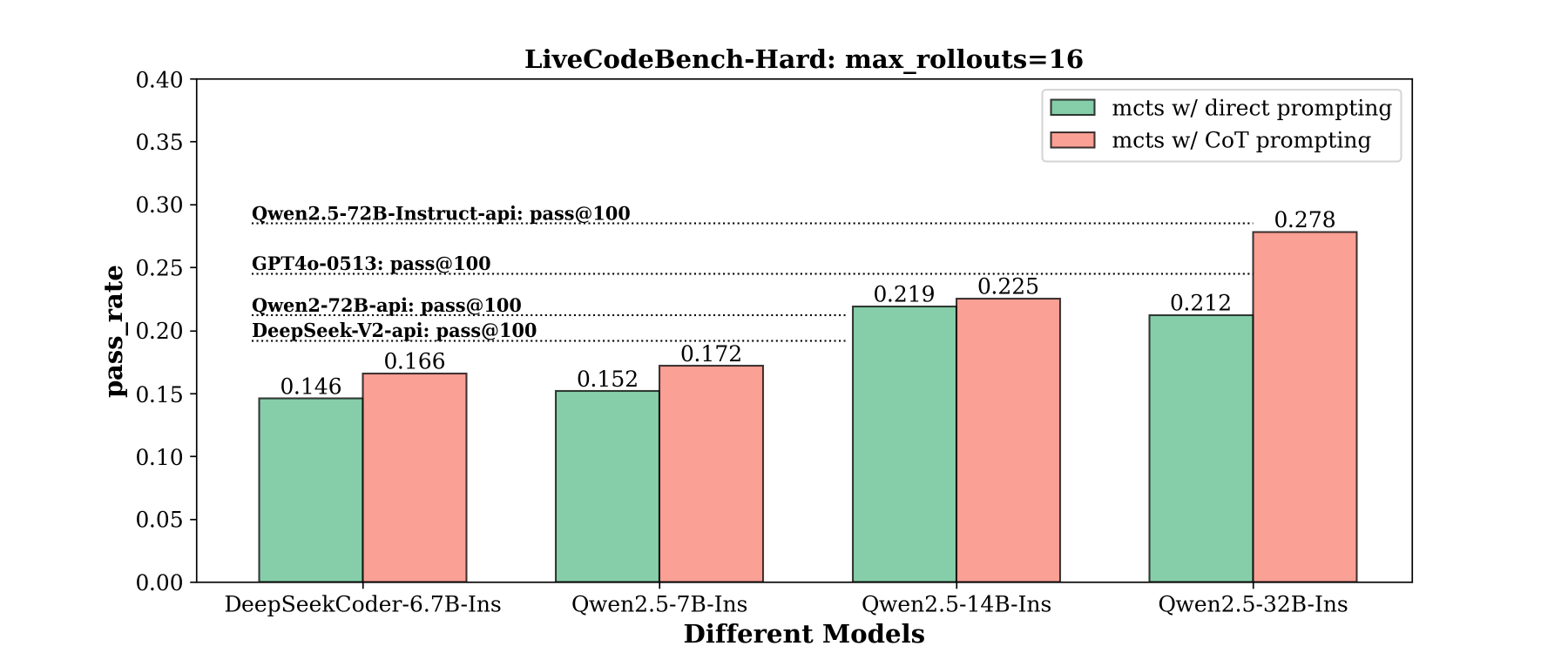}
    \caption{When different models are used as the generating model for MCTS, the pass rates of MCTS tend to increase correspondingly with the enhancement of model capabilities. The pass@100 rates of DeepSeekCoder-6.7B-Instruct, Qwen2.5-7B-Instruct, Qwen2.5-14B-Instruct, and Qwen2.5-32B-Instruct are 0.080, 0.099, 0.189, and 0.197, respectively. It can be observed that after employing MCTS, even with max\_rollouts set to only 16, the performance of each model on LiveCodeBench-Hard significantly exceeds its own pass@100 rate.}
    \label{LiveCodeBench-Hard_Different_Models}
\end{figure}

\begin{figure}[!tbp]
    \centering
    \includegraphics[width=0.8\textwidth]{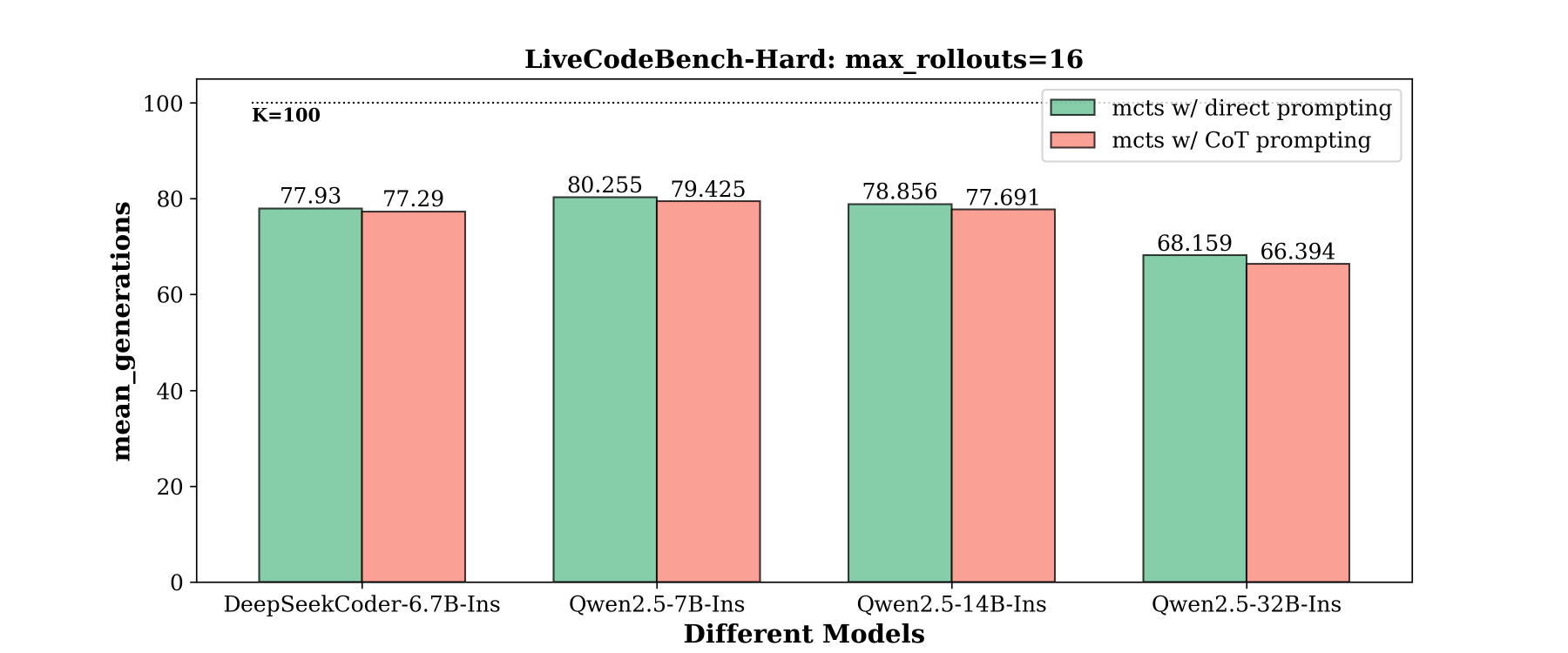}
    \caption{To ensure fairness in comparison with the pass@100 rates, we also recorded the average number of generations produced by MCTS when max\_rollouts is set to 16 across different models. It can be observed that for DeepSeekCoder-6.7B-Instruct, Qwen2.5-7B-Instruct, and Qwen2.5-14B-Instruct, the average number of generations is approximately 80, which is significantly lower than 100. For Qwen2.5-32B-Instruct, the average number of generations is even lower.}
    \label{LiveCodeBench-Hard_Different_Models_mean_generations}
\end{figure}

Furthermore, Figure~\ref{LiveCodeBench-Hard-sota} presents a comparison of MCTS with smaller models against the pass@100 rates of state-of-the-art models. Key findings include:
\begin{itemize}
    \item When $\text{max\_rollouts}=32$, MCTS with DeepSeekCoder-6.7B-Instruct achieves a pass rate of 0.205, which is close to Qwen2.5-72B-Instruct-api's pass rate of 0.212, representing a significant relative improvement of 156.25\% compared to the standalone pass@100 rate of DeepSeekCoder-6.7B-Instruct at 0.080.
    \item Similarly, when $\text{max\_rollouts}=32$, MCTS with Qwen2.5-32B-Instruct achieves a pass rate of 0.278, approaching the performance of Qwen2.5-72B-Instruct-api at 0.285.
\end{itemize}
% When $\text{max\_rollouts}=32$, MCTS with DeepSeekCoder-6.7B-Instruct achieves a pass@100 score of 20.5, close to Qwen2.5-72B-Instruct-api's score of 21.2. This represents a significant relative improvement of 156.25\% compared to the standalone pass@100 score of DeepSeekCoder-6.7B-Instruct at 8.0. Similarly, when $\text{max\_rollouts}=32$, MCTS with Qwen2.5-32B-Instruct achieves a pass@100 score of 27.8, approaching the score of Qwen2.5-72B-Instruct-api at 28.5.

For a comparison under the same number of generations, Figure~\ref{LiveCodeBench-Hard-gen} provides insights into the efficiency of MCTS, with the x-axis representing the mean number of generations. These results highlight that even on this challenging subset, MCTS achieves significantly improved pass rates with fewer sampling attempts. In summary, MCTS demonstrates robust performance on the LiveCodeBench-Hard subset, achieving competitive results against larger models and maintaining efficiency despite the increased difficulty.

The observed improvement of MCTS on LiveCodeBench-Hard is notably greater than that on LiveCodeBench-Medium, indicating that MCTS retains its effectiveness even under increased task difficulty without experiencing performance degradation.

\begin{figure}[!tbp]
    \centering
    \includegraphics[width=0.8\textwidth]{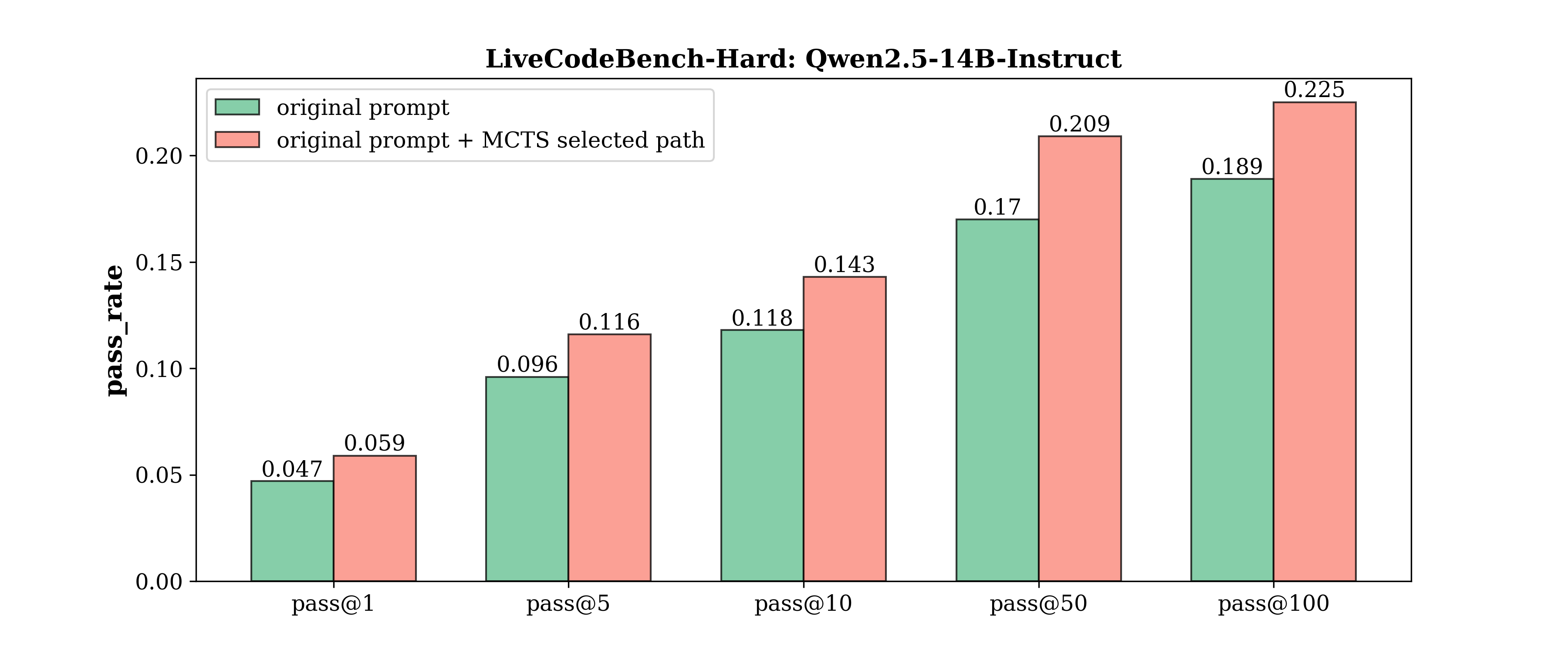}
    \caption{A comparison between the original prompts from LiveCodeBench-Hard and the modified prompts using the optimal paths identified through MCTS. The results show substantial contribution of the paths identified during the selection and expansion phases to the overall performance of MCTS.}
    \label{LiveCodeBench-Hard_MCTS_select}
\end{figure}

\noindent{\bf Model-Agnostic.} To evaluate the model-agnostic nature of the MCTS method, we compare its performance on LiveCodeBench-Hard subset using four different generating models: DeepSeekCoder-6.7B-Instruct, Qwen2.5-7B-Instruct, Qwen2.5-14B-Instruct, and Qwen2.5-32B-Instruct. For experimental efficiency, we fix $\text{max\_rollouts} = 16$.

Figure~\ref{LiveCodeBench-Hard_Different_Models} illustrates that the performance of MCTS improves consistently with the capabilities of the generating models. Notably, when using Qwen2.5-14B-Instruct, MCTS achieves a pass@100 rate of 0.219, surpassing the pass rate of the larger Qwen2.5-72B-api at 0.212.

Additionally, as shown in Figure~\ref{LiveCodeBench-Hard_Different_Models_mean_generations}, the average number of generations for MCTS remains below 81 across all generating models when $\text{max\_rollouts}=16$. These results indicate that the MCTS method is both model-agnostic and efficient, achieving competitive performance across a variety of generating models while maintaining low sampling overhead. It is also important to note that we plan to optimize pruning techniques for MCTS in the future, which may further reduce the average number of generations.

\subsection{Direct Prompting vs CoT Prompting}
Chain-of-Thought (CoT) prompting has proven to be an effective technique for enhancing model performance in reasoning tasks. Throughout our experiments, we also explore whether integrating CoT prompting with MCTS can further improve overall performance.

On LiveCodeBench-Medium, Figure~\ref{LiveCodeBench-Medium} demonstrates that MCTS with CoT prompting consistently outperforms the baseline. For example, with $\text{max\_rollouts} = 64$ and Qwen2.5-Coder-32B-Instruct, MCTS with CoT prompting achieves a pass rate of 0.810, compared to 0.770 with direct prompting. A similar trend is observed on LiveCodeBench-Hard, as shown in Figure~\ref{LiveCodeBench-Hard}: with $\text{max\_rollouts} = 64$ and Qwen2.5-32B-Instruct, MCTS with CoT prompting achieves a pass rate of 0.351, outperforming the pass rate of 0.305 from direct prompting.

Figure~\ref{LiveCodeBench-Hard_Different_Models_mean_generations} further highlights the efficiency of CoT prompting, requiring fewer generations on both datasets when combined with MCTS. Additionally, Figure~\ref{LiveCodeBench-Hard_Different_Models} demonstrates consistent performance improvements across different models when CoT prompting is applied. As model capabilities increase, the combined approach of MCTS with CoT prompting becomes even more effective. To sum up, CoT prompting outperforms pure MCTS across all four tested models, which highlights its robustness and versatility.

\subsection{Deep Insight into MCTS's Selection Phase}
In the context of MCTS, the final generated response can be decomposed into three key components: (1) the path identified during the selection phase via P-UCB search, (2) the actions sampled by the model during the expansion phase, and (3) the content generated through autoregressive decoding in the simulation phase. Since the sampling methods used in the simulation phase are indistinguishable from techniques like Best-of-N, we are particularly interested in the specific effects of the paths identified during the selection and expansion phases. To explore this, we present an interpretive experiment in the sequel.

Using the LiveCodeBench-Hard dataset with Qwen2.5-14B-Instruct, we fix $\text{max\_rollouts} = 16$ and record the best paths discovered by MCTS. We the modify the prompt for each problem to include the format prompt + best path and employ standard autoregressive decoding methods to sample and compute pass@k rates. Figure~\ref{LiveCodeBench-Hard_MCTS_select} compares the pass@k rates of Qwen2.5-14B-Instruct on LiveCodeBench-Hard when using the original prompt versus the modified prompt.

The results demonstrate significant improvements with the modified prompt across all evaluated k-values ($K=1, 5, 10, 50, \text{~and~} 100$). For instance, at $K=1$, the pass rate increases by 25.5\% compared to the baseline, while at K=100, the relative improvement is 19\%. These findings highlight the substantial contribution of the paths identified during the selection and expansion phases to the overall performance of MCTS.

\subsection{CodeContest-Test}
CodeContest-Test is a widely used benchmark set for code competition tasks, complementing LiveCodeBench. To evaluate the generalizability and effectiveness of the methods proposed in this paper across multiple code competition benchmark sets, we also compare our methods against several baseline models on CodeContest-Test.

\begin{figure}[t!]
    \centering
    \includegraphics[width=0.95\textwidth]{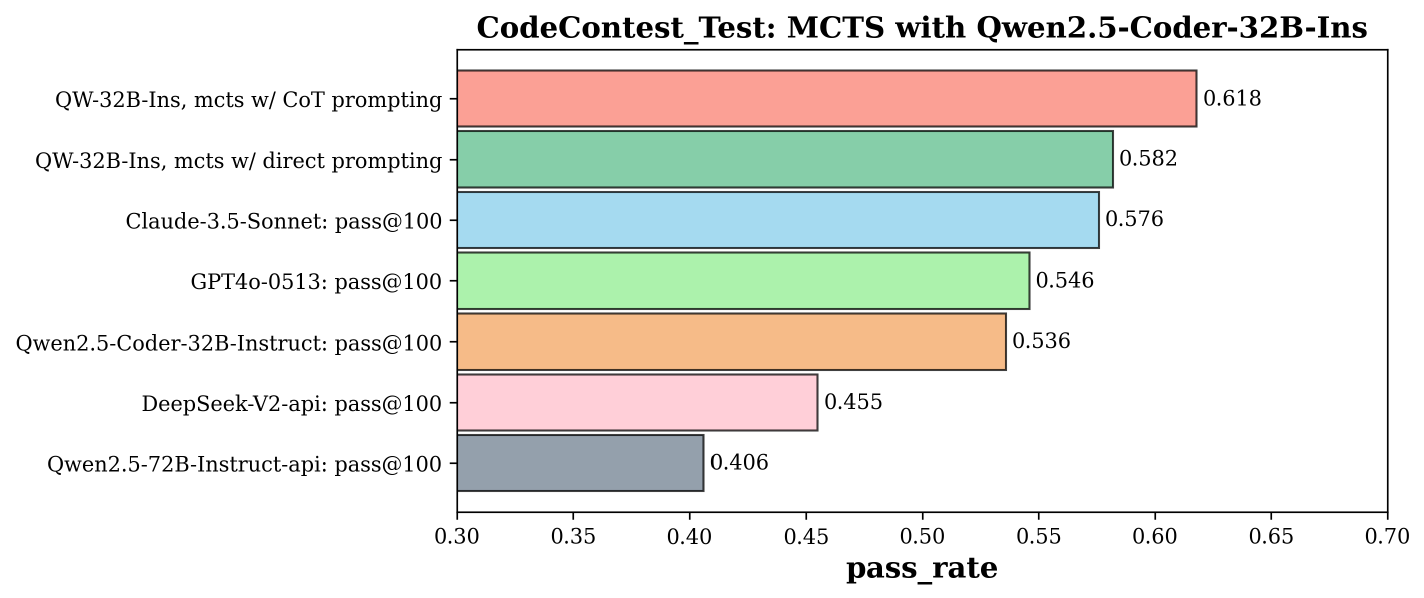}
    \caption{Evaluation results on CodeContest\_Test. MCTS with $\text{max\_rollouts}=16$ and Qwen2.5B-Coder-32B-Instruct, which involves less than 80 number of generations, outperforms pass@100 of Claude-3.5-Sonnet.}
    \label{CodeContest_test}
\end{figure}

To control experimental costs, we utilize the Qwen2.5-Coder-32B-Instruct model and fix the MCTS max\_rollouts at 32. As shown in Figure~\ref{CodeContest_test}, when using direct prompting, MCTS achieves a pass rate of 0.582, surpassing the performance of Claude-3.5-Sonnet. Notably, the average number of generations for MCTS in this setup is only 79.055.

When employing Chain-of-Thought (CoT) prompting, the performance of MCTS improves further, achieving a pass@100 rate of 0.618, while reducing the average number of generations to 75.962. These results demonstrate the effectiveness of combining MCTS with CoT prompting, achieving better performance with fewer sampling attempts.

\section{Conclusions}
In this paper, we proposed a novel token-level Monte Carlo Tree Search (MCTS) framework combined with Chain-of-Thought (CoT) prompting, tailored for competition-level code generation tasks. Using the open-source Qwen2.5-Coder-32B-Instruct model, our approach demonstrates its effectiveness by achieving a pass rate of \textbf{0.351} on LiveCodeBench-Hard, nearing the pass@1 performance of O1-Mini. The results highlight the capability of our framework to significantly improve the problem-solving efficiency and accuracy of open-source models, thereby reducing the reliance on large-scale proprietary black-box LLMs. Moreover, our method's ability to generate consistent and high-quality solutions making it possible to synthesize supervised fine-tuning (SFT) data for large-scale competition level code problems from open-source LLMs. By synthesizing robust datasets directly from the target model, our approach paves the way for more effective and intrinsically aligned post-training strategies.

In the future, our framework can be further enhanced by integrating with techniques like rejection sampling and self-consistent reasoning. These techniques could complement our MCTS framework, further enhancing the LLMs' reasoning capabilities and improving their performance on competition-level code generation tasks. By enabling more robust and diverse exploration of potential solutions, and minimizing the generation of incorrect or incomplete code, these enhancements have the potential to advance the state of the art in solving complex coding problems.

\section{Contributions \& Acknowledgements}
\noindent\textbullet\ \textbf{Paper Writing:} Hao Wang, Boyi Liu, Jie Chen

\noindent\textbullet\ \textbf{Evaluation:} Hao Wang, Jie Chen

\noindent\textbullet\ \textbf{MCTS Implementation:} Hao Wang, Boyi Liu, Yufeng Zhang, Jie Chen

\noindent\textbullet\ \textbf{Model Serving:} Yufeng Zhang, Hao Wang

\noindent\textbullet\ \textbf{Code Sandbox Integration:} Hao Wang, Jie Chen

We thank Qi Liu for his helpful discussion in model evaluations. We thank Siyao Liu for providing the sandbox environment. We thank Xierui Song for his helpful feedback in the development of MCTS. Their commitment and expertise are crucial to the success of this project.

% \bibliography{reference}

\begin{thebibliography}{99}

\bibitem[Chen et al.(2021)]{chen2021evaluating}
Chen, M., Tworek, J., Jun, H., Yuan, Q., Pinto, H. P. D. O., Kaplan, J., Edwards, H., Burda, Y., Joseph, N., Brockman, G., et al. (2021). Evaluating large language models trained on code. \textit{arXiv preprint arXiv:2107.03374}.

\bibitem[Brown(2020)]{brown2020language}
Brown, T. B. (2020). Language models are few-shot learners. \textit{arXiv preprint arXiv:2005.14165}.

\bibitem[Luo et al.(2023)]{luo2023wizardcoder}
Luo, Z., Xu, C., Zhao, P., Sun, Q., Geng, X., Hu, W., Tao, C., Ma, J., Lin, Q., Jiang, D. (2023). Wizardcoder: Empowering code large language models with evol-instruct. \textit{arXiv preprint arXiv:2306.08568}.

\bibitem[Zheng et al.(2023)]{zheng2023codegeex}
Zheng, Q., Xia, X., Zou, X., Dong, Y., Wang, S., Xue, Y., Shen, L., Wang, Z., Wang, A., Li, Y., et al. (2023). Codegeex: A pre-trained model for code generation with multilingual benchmarking on humaneval-x. In \textit{Proceedings of the 29th ACM SIGKDD Conference on Knowledge Discovery and Data Mining} (pp. 5673--5684).

\bibitem[Li et al.(2022)]{li2022competition}
Li, Y., Choi, D., Chung, J., Kushman, N., Schrittwieser, J., Leblond, R., Eccles, T., Keeling, J., Gimeno, F., Dal Lago, A., et al. (2022). Competition-level code generation with alphacode. \textit{Science}, 378(6624), 1092--1097.

\bibitem[Guo et al.(2024)]{guo2024deepseek}
Guo, D., Zhu, Q., Yang, D., Xie, Z., Dong, K., Zhang, W., Chen, G., Bi, X., Wu, Y., Li, Y. K., et al. (2024). DeepSeek-Coder: When the Large Language Model Meets Programming--The Rise of Code Intelligence. \textit{arXiv preprint arXiv:2401.14196}.

\bibitem[Hui et al.(2024)]{hui2024qwen2}
Hui, B., Yang, J., Cui, Z., Yang, J., Liu, D., Zhang, L., Liu, T., Zhang, J., Yu, B., Lu, K., et al. (2024). Qwen2. 5-coder technical report. \textit{arXiv preprint arXiv:2409.12186}.

\bibitem[Li et al.(2023)]{li2023starcoder}
Li, R., Allal, L. B., Zi, Y., Muennighoff, N., Kocetkov, D., Mou, C., Marone, M., Akiki, C., Li, J., Chim, J., et al. (2023). Starcoder: may the source be with you!. \textit{arXiv preprint arXiv:2305.06161}.

\bibitem[Roziere et al.(2023)]{roziere2023code}
Roziere, B., Gehring, J., Gloeckle, F., Sootla, S., Gat, I., Tan, X. E., Adi, Y., Liu, J., Sauvestre, R., Remez, T., et al. (2023). Code llama: Open foundation models for code. \textit{arXiv preprint arXiv:2308.12950}.

\bibitem[Achiam et al.(2023)]{achiam2023gpt}
Achiam, J., Adler, S., Agarwal, S., Ahmad, L., Akkaya, I., Aleman, F. L., Almeida, D., Altenschmidt, J., Altman, S., Anadkat, S., et al. (2023). Gpt-4 technical report. \textit{arXiv preprint arXiv:2303.08774}.

\bibitem[Shinn et al.(2024)]{shinn2024reflexion}
Shinn, N., Cassano, F., Gopinath, A., Narasimhan, K., Yao, S. (2024). Reflexion: Language agents with verbal reinforcement learning. \textit{Advances in Neural Information Processing Systems}, 36.

\bibitem[Yao et al.(2022)]{yao2022react}
Yao, S., Zhao, J., Yu, D., Du, N., Shafran, I., Narasimhan, K., Cao, Y. (2022). React: Synergizing reasoning and acting in language models. \textit{arXiv preprint arXiv:2210.03629}.

\bibitem[Huang et al.(2023)]{huang2023agentcoder}
Huang, D., Bu, Q., Zhang, J. M., Luck, M., Cui, H. (2023). Agentcoder: Multi-agent-based code generation with iterative testing and optimisation. \textit{arXiv preprint arXiv:2312.13010}.

\bibitem[Hong et al.(2023)]{hong2023metagpt}
Hong, S., Zheng, X., Chen, J., Cheng, Y., Wang, J., Zhang, C., Wang, Z., Yau, S. K. S., Lin, Z., Zhou, L., et al. (2023). Metagpt: Meta programming for multi-agent collaborative framework. \textit{arXiv preprint arXiv:2308.00352}.

\bibitem[Wang et al.(2024)]{wang2024executable}
Wang, X., Chen, Y., Yuan, L., Zhang, Y., Li, Y., Peng, H., Ji, H. (2024). Executable code actions elicit better llm agents. \textit{arXiv preprint arXiv:2402.01030}.

\bibitem[Zhang et al.(2024)]{zhang2024autocoderover}
Zhang, Y., Ruan, H., Fan, Z., Roychoudhury, A. (2024). Autocoderover: Autonomous program improvement. In \textit{Proceedings of the 33rd ACM SIGSOFT International Symposium on Software Testing and Analysis} (pp. 1592--1604).

\bibitem[Zhang et al.(2023)]{zhang2023planning}
Zhang, S., Chen, Z., Shen, Y., Ding, M., Tenenbaum, J. B., Gan, C. (2023). Planning with large language models for code generation. \textit{arXiv preprint arXiv:2303.05510}.

\bibitem[Wei et al.(2022)]{wei2022chain}
Wei, J., Wang, X., Schuurmans, D., Bosma, M., Xia, F., Chi, E., Le, Q. V., Zhou, D., et al. (2022). Chain-of-thought prompting elicits reasoning in large language models. \textit{Advances in neural information processing systems}, 35, 24824--24837.

\bibitem[Wang et al.(2022)]{wang2022self}
Wang, X., Wei, J., Schuurmans, D., Le, Q., Chi, E., Narang, S., Chowdhery, A., Zhou, D. (2022). Self-consistency improves chain of thought reasoning in language models. \textit{arXiv preprint arXiv:2203.11171}.

\bibitem[Yao et al.(2024)]{yao2024tree}
Yao, S., Yu, D., Zhao, J., Shafran, I., Griffiths, T., Cao, Y., Narasimhan, K. (2024). Tree of thoughts: Deliberate problem solving with large language models. \textit{Advances in Neural Information Processing Systems}, 36.

\bibitem[Hao et al.(2023)]{hao2023reasoning}
Hao, S., Gu, Y., Ma, H., Hong, J. J., Wang, Z., Wang, D. Z., Hu, Z. (2023). Reasoning with language model is planning with world model. \textit{arXiv preprint arXiv:2305.14992}.

\bibitem[Madaan et al.(2024)]{madaan2024self}
Madaan, A., Tandon, N., Gupta, P., Hallinan, S., Gao, L., Wiegreffe, S., Alon, U., Dziri, N., Prabhumoye, S., Yang, Y., et al. (2024). Self-refine: Iterative refinement with self-feedback. \textit{Advances in Neural Information Processing Systems}, 36.

\bibitem[Zhou et al.(2023)]{zhou2023language}
Zhou, A., Yan, K., Shlapentokh-Rothman, M., Wang, H., Wang, Y. X. (2023). Language agent tree search unifies reasoning acting and planning in language models. \textit{arXiv preprint arXiv:2310.04406}.

\bibitem[Chen et al.(2023)]{chen2023teaching}
Chen, X., Lin, M., Schärli, N., Zhou, D. (2023). Teaching large language models to self-debug. \textit{arXiv preprint arXiv:2304.05128}.

\bibitem[Huang et al.(2023)]{huang2023codecot}
Huang, D., Bu, Q., Cui, H. (2023). Codecot and beyond: Learning to program and test like a developer. \textit{arXiv preprint arXiv:2308.08784}.

\bibitem[Zhang et al.(2024)]{zhang2024rest}
Zhang, D., Zhoubian, S., Hu, Z., Yue, Y., Dong, Y., Tang, J. (2024). Rest-mcts*: Llm self-training via process reward guided tree search. \textit{arXiv preprint arXiv:2406.03816}.

\bibitem[Li et al.(2024)]{li2024rethinkmcts}
Li, Q., Xia, W., Du, K., Dai, X., Tang, R., Wang, Y., Yu, Y., Zhang, W. (2024). RethinkMCTS: Refining Erroneous Thoughts in Monte Carlo Tree Search for Code Generation. \textit{arXiv preprint arXiv:2409.09584}.

\bibitem[Qi et al.(2024)]{qi2024mutual}
Qi, Z., Ma, M., Xu, J., Zhang, L. L., Yang, F., Yang, M. (2024). Mutual reasoning makes smaller llms stronger problem-solvers. \textit{arXiv preprint arXiv:2408.06195}.

\bibitem[Wang et al.(2024)]{wang2024q}
Wang, C., Deng, Y., Lyu, Z., Zeng, L., He, J., Yan, S., An, B. (2024). Q*: Improving multi-step reasoning for llms with deliberative planning. \textit{arXiv preprint arXiv:2406.14283}.

\bibitem[Hui et al.(2024)]{hui2024rot}
Hui, W., Wang, Y., Tu, K., Jiang, C. (2024). RoT: Enhancing Large Language Models with Reflection on Search Trees. \textit{arXiv preprint arXiv:2404.05449}.

\bibitem[Jiang et al.(2024)]{jiang2024self}
Jiang, X., Dong, Y., Wang, L., Fang, Z., Shang, Q., Li, G., Jin, Z., Jiao, W. (2024). Self-planning code generation with large language models. \textit{ACM Transactions on Software Engineering and Methodology}, 33(7), 1--30.

\bibitem[Wang et al.(2024)]{wang2024planning}
Wang, E., Cassano, F., Wu, C., Bai, Y., Song, W., Nath, V., Han, Z., Hendryx, S., Yue, S., Zhang, H. (2024). Planning in natural language improves llm search for code generation. \textit{arXiv preprint arXiv:2409.03733}.

\bibitem[Silver et al.(2017)]{silver2017mastering}
Silver, D., Hubert, T., Schrittwieser, J., Antonoglou, I., Lai, M., Guez, A., Lanctot, M., Sifre, L., Kumaran, D., Graepel, T., et al. (2017). Mastering chess and shogi by self-play with a general reinforcement learning algorithm. \textit{arXiv preprint arXiv:1712.01815}.

\bibitem[Silver et al.(2016)]{silver2016mastering}
Silver, D., Huang, A., Maddison, C. J., Guez, A., Sifre, L., Van Den Driessche, G., Schrittwieser, J., Antonoglou, I., Panneershelvam, V., Lanctot, M., et al. (2016). Mastering the game of Go with deep neural networks and tree search. \textit{Nature}, 529(7587), 484--489.

\bibitem[Naman et al.(2024)]{naman2024livecodebench}
Naman Jain, K. H., Gu, A., Li, W.-D., Yan, F., Zhang, T., Wang, S., Solar-Lezama, A., Sen, K., Stoica, I. (2024). Livecodebench: Holistic and contamination free evaluation of large language models for code. \textit{arXiv preprint arXiv:2403.07974}.

\bibitem[Shinn et al.(2023)]{shinn2023reflexion}
Shinn, N., Cassano, F., Labash, B., Gopinath, A., Narasimhan, K., Yao, S. (2023). Reflexion: Language agents with verbal reinforcement learning. \textit{arXiv preprint cs.AI/2303.11366}.

\end{thebibliography}

%%%%%%%%%%%%%%%%%%%%%%%%%%%%%%%%%%%%%%%%%%%%%%%%%%%%%%%%%%%%

\newpage
\appendix

% \section{Appendix}

% \subsection{Comparison between MCTS and Best-of-N}
% \begin{figure}[h]
%   \centering
%     \subfigure[]{
%         \includegraphics[width=0.95\textwidth]{figures/LiveCodeBench-Medium_DS6B7_3.png}
%         \label{LiveCodeBench-Medium_bon}
%     }
%     \subfigure[]{
%         \includegraphics[width=0.95\textwidth]{figures/LiveCodeBench-Hard_DS6B7_3.png}
%         \label{LiveCodeBench-Hard_bon}
%     }
%     \caption{Comparison between MCTS and Best-of-N. (a) LiveCodeBench-Medium. (b) LiveCodeBench-Hard.}
%     \label{LiveCodeBench-bon}
% \end{figure}

\section{Extra Tables of MCTS and Pass@k on LiveCodeBench}\label{appendix:table}

\subsection{LiveCodeBench-Medium}
% As shown in Table~\ref{baseline-MCTS-LiveCodeBench-Medium}, the mean number of generations of MCTS with DeepSeekCoder-6.7B-Instruct is approximately 93 when $\text{max\_rollouts} = 32$, and for Qwen2.5-32B-Instruct, it is around 80 when $\text{max\_rollouts} = 64$. Neither exceeds 100 generations, indicating that the proposed approach achieves higher scores with fewer sampling attempts.

\begin{table}[H]
\caption{Evaluation results on LiveCodeBench-Medium: pass@k}
\label{baseline-pass@k-LiveCodeBench-Medium}
\begin{tabular}{|lccccc|}
\hline
\multicolumn{1}{|c|}{\textbf{Model}}                               & \textbf{pass@1} & \textbf{pass@5} & \textbf{pass@10} & \textbf{pass@50} & \textbf{pass@100} \\ \hline
\multicolumn{1}{|l|}{\textbf{Qwen2.5-72B-Instruct-api}}            & 0.477           & 0.611           & 0.646            & 0.716            & \colorbox{blue!20}{0.743} \\
\multicolumn{1}{|l|}{\textbf{Qwen2.5-Coder-32B-Instruct}}          & 0.300           & 0.511           & 0.575            & 0.657            & 0.683             \\
\multicolumn{1}{|l|}{\textbf{GPT4o-0513}}                          & 0.387           & 0.528           & 0.567            & 0.638            & 0.665             \\
\multicolumn{1}{|l|}{\textbf{Qwen2-72B-api}}                       & 0.214           & 0.394           & 0.461            & 0.589            & 0.637             \\
\multicolumn{1}{|l|}{\textbf{Qwen2.5-Coder-7B-Instruct}}           & 0.275           & 0.444           & 0.506            & 0.602            & 0.625             \\
\multicolumn{1}{|l|}{\textbf{DeepSeek-V2-api}}                     & 0.432           & 0.521           & 0.550            & 0.592            & 0.604             \\
\multicolumn{1}{|l|}{\textbf{GLM-4-0520}}                          & 0.190           & 0.336           & 0.398            & 0.510            & 0.551             \\
\multicolumn{1}{|l|}{\textbf{GPT4o-mini-0718}}                     & 0.292           & 0.427           & 0.468            & 0.524            & 0.543             \\
\multicolumn{1}{|l|}{\textbf{Gemini-1.5-pro}}                      & 0.210           & 0.314           & 0.359            & 0.462            & 0.502             \\
\multicolumn{1}{|l|}{\textbf{DeepSeekCoder-6.7B-Instruct}}         & 0.088           & 0.194           & 0.240            & 0.319            & 0.355             \\ \hline
\end{tabular}
\end{table}

\begin{table}[H]
\caption{Evaluation results on LiveCodeBench-Medium: MCTS Results}
\label{baseline-MCTS-LiveCodeBench-Medium}
\begin{tabular}{|l|ccccc|}
\hline
\textbf{Metric \textbackslash Rollouts}       & \textbf{4}   & \textbf{8}   & \textbf{16}  & \textbf{32}  & \textbf{64}   \\ \hline
\multicolumn{6}{|c|}{\textbf{DeepSeekCoder-6.7B-Instruct}}                                                                \\ \hline
pass rate w/direct prompting               & 0.270 & 0.325 & 0.392 & 0.488 & 0.502 \\
pass rate w/CoT prompting                  & 0.306 & 0.347 & 0.433 & 0.502 & 0.531 \\
mean generations w/direct prompting        & 15.990 & 29.547 & 54.824 & 93.024 & 182.572 \\
mean generations w/CoT prompting           & 14.384 & 27.871 & 51.063 & 92.519 & 176.505 \\ \hline
\multicolumn{6}{|c|}{\textbf{Qwen2.5-Coder-32B-Instruct}}                                                                 \\ \hline
pass rate w/direct prompting               & 0.648 & 0.676 & 0.720 & 0.743 & 0.770 \\
pass rate w/CoT prompting                  & 0.686 & 0.729 & 0.755 & 0.770 & \colorbox{blue!20}{0.810} \\
mean generations w/direct prompting        & 9.871 & 15.943 & 24.053 & 47.089 & 79.751 \\
mean generations w/CoT prompting           & 8.455 & 14.360 & 22.579 & 45.163 & 70.767 \\ \hline
\end{tabular}
\end{table}

\subsection{LiveCodeBench-Hard}

\begin{table}[H]
\caption{Evaluation results on LiveCodeBench-Hard: pass@k Results}
\label{baseline-LiveCodeBench-Hard-passk}
\begin{tabular}{|lccccc|}
\hline
\textbf{Model}                              & \textbf{pass@1} & \textbf{pass@5} & \textbf{pass@10} & \textbf{pass@50} & \textbf{pass@100} \\ \hline
\textbf{Qwen2.5-72B-Instruct-api}           & 0.087           & 0.150           & 0.182            & 0.256            & \colorbox{blue!20}{0.285} \\
\textbf{GPT4o-0513}                         & 0.068           & 0.133           & 0.161            & 0.223            & 0.245 \\
\textbf{Qwen2-72B-api}                      & 0.025           & 0.070           & 0.094            & 0.168            & 0.212 \\
\textbf{Qwen2.5-Coder-32B-Instruct}         & 0.054           & 0.102           & 0.123            & 0.174            & 0.197 \\
\textbf{DeepSeek-V2-api}                    & 0.090           & 0.138           & 0.156            & 0.187            & 0.192 \\
\textbf{Gemini-1.5-pro}                     & 0.035           & 0.068           & 0.088            & 0.136            & 0.159 \\
\textbf{GLM-4-0520}                         & 0.013           & 0.035           & 0.053            & 0.105            & 0.133 \\
\textbf{GPT4o-mini-0718}                    & 0.043           & 0.068           & 0.078            & 0.107            & 0.119 \\
\textbf{Qwen2.5-Coder-7B-Instruct}          & 0.025           & 0.050           & 0.061            & 0.083            & 0.099 \\
\textbf{DeepSeekCoder-6.7B-Instruct}        & 0.004           & 0.016           & 0.025            & 0.062            & 0.080 \\ \hline
\end{tabular}
\end{table}

\begin{table}[H]
\caption{Evaluation results on LiveCodeBench-Hard: MCTS Results}
\label{baseline-MCTS-LiveCodeBench-Hard}
\begin{tabular}{|l|ccccc|}
\hline
\textbf{Model \textbackslash Metric}        & \textbf{4}   & \textbf{8}   & \textbf{16}   & \textbf{32}  & \textbf{64} \\ \hline
\multicolumn{6}{|c|}{\textbf{DeepSeekCoder-6.7B-Instruct}}                                                              \\ \hline
pass rate w/direct prompting                        & 0.040        & 0.086        & 0.146        & 0.205         & 0.212         \\
pass rate w/CoT prompting                           & 0.083        & 0.102        & 0.166        & 0.217         & 0.232         \\
mean generations w/direct prompting                 & 17.721       & 37.513       & 77.932       & 133.946       & 289.51         \\
mean generations w/CoT prompting                    & 17.635       & 36.933       & 77.297       & 133.471       & 285.396         \\ \hline
\multicolumn{6}{|c|}{\textbf{Qwen2.5-Coder-32B-Instruct}}                                                              \\ \hline
pass rate w/direct prompting                            & 0.145        & 0.179        & 0.212        & 0.278        &0.305   \\
pass rate w/CoT prompting                               & 0.193        & 0.257        & 0.278        & 0.311        &\colorbox{blue!20}{0.351} \\
mean generations w/direct prompting                     & 18.359       & 28.519       & 68.159       & 121.616        &241.927 \\
mean generations w/CoT prompting                        & 17.303       & 32.912       & 66.394       & 115.072        &228.696 \\ \hline
\end{tabular}
\end{table}

\section{Case Comparison on Best-of-100 and our method}

Here is an example where our method passes all the test cases, but the Best-of-100 method does not:

\lstset{
    backgroundcolor=\color{white},   % 背景色
    basicstyle=\ttfamily\footnotesize, % 字体及大小
    breaklines=true,                  % 自动换行
    captionpos=b,                     % 标题位置
    commentstyle=\color{gray},         % 注释颜色
    keywordstyle=\color{blue},         % 关键字颜色
    stringstyle=\color{red},           % 字符串颜色
    numbers=left,                     % 行号
    numberstyle=\tiny\color{gray},    % 行号样式
    frame=single,                     % 边框
    rulecolor=\color{black}           % 边框颜色
}

\subsection*{Problem Description of Leetcode/2839}

You are given two 0-indexed integer arrays \texttt{nums1} and \texttt{nums2}, each of length \texttt{n}, and a 1-indexed 2D array \texttt{queries} where \texttt{queries[i] = [x\_i, y\_i]}. For the $i^{th}$ query, find the maximum value of \texttt{nums1[j] + nums2[j]} among all indices \texttt{j} $(0 \leq j < n)$, where \texttt{nums1[j] >= x\_i} and \texttt{nums2[j] >= y\_i}, or $-1$ if there is no \texttt{j} satisfying the constraints.

Return an array \texttt{answer} where \texttt{answer[i]} is the answer to the $i^{th}$ query.

\subsection*{Constraints}
\begin{itemize}
    \item \texttt{nums1.length == nums2.length}
    \item \texttt{1 <= n <= $10^5$}
    \item \texttt{1 <= nums1[i], nums2[i] <= $10^9$}
    \item \texttt{1 <= queries.length <= $10^5$}
    \item \texttt{1 <= x\_i, y\_i <= $10^9$}
\end{itemize}

\subsection*{Examples}
\begin{itemize}
    \item \textbf{Example 1:}\\
    Input: nums1 = [4,3,1,2], nums2 = [2,4,9,5], queries = [[4,1],[1,3],[2,5]]\\
    Output: [6,10,7]
    \item \textbf{Example 2:}\\
    Input: nums1 = [3,2,5], nums2 = [2,3,4], queries = [[4,4],[3,2],[1,1]]\\
    Output: [9,9,9]
    \item \textbf{Example 3:}\\
    Input: nums1 = [2,1], nums2 = [2,3], queries = [[3,3]]\\
    Output: [-1]
\end{itemize}

\subsection*{Best-of-100 on Leetcode/2839}

\begin{lstlisting}[language=Python]
class Solution:
    def maximumSumQueries(self, nums1: List[int], nums2: List[int], queries: List[List[int]]) -> List[int]:
        from sortedcontainers import SortedList
        
        # Combine nums1 and nums2 into pairs and sort them by nums1 in descending order
        combined = sorted(zip(nums1, nums2), reverse=True)
        query_with_index = [(x, y, i) for i, (x, y) in enumerate(queries)]
        query_with_index.sort(reverse=True)
        
        # Initialize a sorted list to keep track of the maximum sums
        max_sums = SortedList()
        result = [-1] * len(queries)
        j = 0
        
        for x, y, i in query_with_index:
            # Add valid pairs to max_sums
            while j < len(combined) and combined[j][0] >= x:
                num1, num2 = combined[j]
                # Maintain the sorted list with only the maximum sums for each unique num2
                if not max_sums or max_sums[-1][1] < num1 + num2:
                    max_sums.add((num2, num1 + num2))
                j += 1
            
            # Find the maximum sum for the current query
            idx = max_sums.bisect_left((y, float('inf')))
            if idx > 0:
                result[i] = max_sums[idx-1][1]
        
        return result
\end{lstlisting}

\subsection*{Our method on Leetcode/2839}

\begin{lstlisting}[language=Python]
from typing import List

class Solution:
    def maximumSumQueries(self, nums1: List[int], nums2: List[int], queries: List[List[int]]) -> List[int]:
        combined = sorted(zip(nums1, nums2), reverse=True)
        queries = sorted([(x, y, i) for i, (x, y) in enumerate(queries)], reverse=True)
        result = [-1] * len(queries)
        stack = []
        
        for x, y, i in queries:
            while combined and combined[0][0] >= x:
                a, b = combined.pop(0)
                while stack and stack[-1][0] <= a + b:
                    stack.pop()
                stack.append((a + b, b))
            for val, min_b in stack:
                if min_b >= y:
                    result[i] = val
                    break
        
        return result
\end{lstlisting}

\lstset{
    backgroundcolor=\color{white},   % 背景色
    basicstyle=\ttfamily\footnotesize, % 字体及大小
    breaklines=true,                  % 自动换行
    captionpos=b,                     % 标题位置
    commentstyle=\color{gray},         % 注释颜色
    keywordstyle=\color{blue},         % 关键字颜色
    stringstyle=\color{red},           % 字符串颜色
    numbers=left,                     % 行号
    numberstyle=\tiny\color{gray},    % 行号样式
    frame=single,                     % 边框
    rulecolor=\color{black}           % 边框颜色
}

\subsection*{Problem Description of atcoder/abc322\_e}

AtCoder Inc. is planning to develop a product. The product has $K$ parameters, whose values are currently all zero. The company aims to raise all parameter values to at least $P$. 

There are $N$ development plans. Executing the $i^{th}$ development plan $(1 \leq i \leq N)$ increases the value of the $j^{th}$ parameter by $A_{i,j}$ for every integer $j$ such that $1 \leq j \leq K$, at the cost of $C_i$. A development plan cannot be executed more than once. 

Determine whether the company can achieve its goal, and if it can, find the minimum total cost required to achieve the goal.

\subsection*{Constraints}
\begin{itemize}
    \item $1 \leq N \leq 100$
    \item $1 \leq K, P \leq 5$
    \item $0 \leq A_{i,j} \leq P \quad (1 \leq i \leq N, 1 \leq j \leq K)$
    \item $1 \leq C_i \leq 10^9 \quad (1 \leq i \leq N)$
    \item All input values are integers.
\end{itemize}

\subsection*{Examples}
\begin{itemize}
    \item \textbf{Example 1:}\\
    Input:
    \begin{verbatim}
4 3 5
5 3 0 2
3 1 2 3
3 2 4 0
1 0 1 4
    \end{verbatim}
    Output:
    \begin{verbatim}
9
    \end{verbatim}
    Explanation: If you execute the first, third, and fourth development plans, each parameter will be $3+2+0=5$, $0+4+1=5$, $2+0+4=6$, all of which are at least $5$, so the goal is achieved. The total cost in this case is $5 + 3 + 1 = 9$. It is impossible to achieve the goal at a total cost of $8$ or less. Thus, the answer is $9$.

    \item \textbf{Example 2:}\\
    Input:
    \begin{verbatim}
7 3 5
85 1 0 1
37 1 1 0
38 2 0 0
45 0 2 2
67 1 1 0
12 2 2 0
94 2 2 1
    \end{verbatim}
    Output:
    \begin{verbatim}
-1
    \end{verbatim}
    Explanation: You cannot achieve the goal no matter what you do. Thus, print $-1$.
\end{itemize}

\subsection*{Best-of-100 on Atcoder/abc322\_e}

\begin{lstlisting}[language=Python]
import sys
import itertools

def main():
    import sys
    import itertools

    def readints():
        return list(map(int, sys.stdin.readline().split()))

    N, K, P = readints()
    plans = [readints() for _ in range(N)]

    # Generate all possible combinations of plans
    min_cost = float('inf')
    for r in range(1, N + 1):
        for combination in itertools.combinations(range(N), r):
            total_cost = sum(plans[i][0] for i in combination)
            total_increase = [sum(plans[i][j + 1] for i in combination) for j in range(K)]
            if all(increase >= P for increase in total_increase):
                min_cost = min(min_cost, total_cost)

    if min_cost == float('inf'):
        print(-1)
    else:
        print(min_cost)

if __name__ == "__main__":
    main()
\end{lstlisting}

\subsection*{Our method on Atcoder/abc322\_e}

\begin{lstlisting}[language=Python]
# YOUR CODE HERE
import sys
input = sys.stdin.read
data = input().split()

N = int(data[0])
K = int(data[1])
P = int(data[2])

plans = []
index = 3
for _ in range(N):
    C = int(data[index])
    A = list(map(int, data[index + 1:index + 1 + K]))
    plans.append((C, A))
    index += 1 + K

min_cost = float('inf')

def dfs(current_params, current_cost, plan_index):
    global min_cost
    if current_cost >= min_cost:
        return
    if all(param >= P for param in current_params):
        min_cost = current_cost
        return
    if plan_index == N:
        return
    # Include the current plan
    new_params = [a + b for a, b in zip(current_params, plans[plan_index][1])]
    dfs(new_params, current_cost + plans[plan_index][0], plan_index + 1)
    # Exclude the current plan
    dfs(current_params, current_cost, plan_index + 1)

dfs([0] * K, 0, 0)

if min_cost == float('inf'):
    print(-1)
else:
    print(min_cost)
\end{lstlisting}

%%%%%%%%%%%%%%%%%%%%%%%%%%%%%%%%%%%%%%%%%%%%%%%%%%%%%%%%%%%%

\end{document}